\documentclass[journal]{IEEEtran}
\usepackage{cite}
\ifCLASSINFOpdf
   \usepackage[pdftex]{graphicx}
\else
\fi
\usepackage[table]{xcolor}
\usepackage{multirow}

\usepackage[cmex10]{amsmath}
\usepackage{algorithmic}
\usepackage{array}
\begin{document}
\title{Sparse Coding for Alpha Matting}

\author{Jubin~Johnson,
          Ehsan~Shahrian~Varnousfaderani,
          Hisham~Cholakkal,
        and~Deepu~Rajan% <-this % stops a space
\thanks{J. Johnson, H. Cholakkal, and D. Rajan are with the Multimedia Lab, School of Computer Science and
Engineering, Nanyang Technological University, Singapore,
639798 (e-mail: jubin001@e.ntu.edu.sg; hisham002@e.ntu.edu.sg; asdrajan@ntu.edu.sg).}% <-this % stops a space
\thanks{E. S. Varnousfaderani is with the Shiley Eye Center, University of California San Diego e-mail: eshahria@ucsd.edu.}
% <-this % stops a space
%\thanks{Manuscript received May 30, 2015
%; revised September 17, 2014.
%}
}


\markboth{
IEEE Transactions on Image Processing 
%,~Vol.~13, No.~9, September~2014
}%
{Shell \MakeLowercase{\textit{et al.}}: Bare Demo of IEEEtran.cls for Journals}



\maketitle

\begin{abstract}
Existing color sampling based alpha matting methods use the compositing equation to estimate alpha at a pixel from pairs of foreground (\textit{F}) and background (\textit{B}) samples. The quality of the matte depends on the selected (\textit{F,B}) pairs. In this paper, the matting problem is reinterpreted as a sparse coding of pixel features, wherein the sum of the codes gives the estimate of the alpha matte from a set of unpaired \textit{F} and \textit{B} samples. A non-parametric probabilistic segmentation provides a certainty measure on the pixel belonging to foreground or background, based on which a dictionary is formed for use in sparse coding. By removing the restriction to conform to (\textit{F,B}) pairs, this method allows for better alpha estimation from multiple \textit{F} and \textit{B} samples. The same framework is extended to videos, where the requirement of temporal coherence is handled effectively. Here, the dictionary is formed by samples from multiple frames. A multi-frame graph model, as opposed to a single image as for image matting, is proposed that can be solved efficiently in closed form. Quantitative and qualitative evaluations on a benchmark dataset are provided to show that the proposed method outperforms current state-of-the-art in image and video matting.

\end{abstract}


% \fi
\IEEEpeerreviewmaketitle

\section{Introduction}
%\IEEEPARstart{T}{his} demo file is intended to serve as a ``starter file''
%for IEEE journal papers produced under \LaTeX\ using
%IEEEtran.cls version 1.8a and later.
\IEEEPARstart{D}{igital} matting is a useful tool for image and video editing where foreground objects need to be extracted accurately and pasted onto a different background.
The color $I_i$ of a pixel $i$ in an image can be considered to be a composite of a foreground color $F_i$ and a background color $B_i$ such that 
\begin{equation} \label{eq:matting}
I_i=\alpha_i F_i + (1-\alpha_i )B_i ,
\end{equation}
where $\alpha$ defines the opacity of the pixel and is a value in [0, 1], with 0 for background pixels and 1 for foreground pixels. Determining $\alpha$ for every pixel, also called \emph{pulling an alpha matte}, is a highly ill-posed problem since it involves estimation of seven unknowns (3 color components for each of $F_i$ and $B_i$ and the $\alpha$ value) from three equations. The problem is constrained by providing additional information in the form of a three-level segmented image known as a trimap~\cite{shahrian2012weighted,shahrian2013improving,chen2013image}, or as scribbles~\cite{levin2008closed,chen2012knn} specifying the definite foreground ($F$), definite background ($B$) and unknown regions.

There are three main approaches for matting: sampling \cite{chuang2001bayesian,he2011global,gastal2010shared,shahrian2012weighted,shahrian2013improving}, alpha propagation \cite{levin2008closed,chen2012knn,beiiterative2013,shi2013color} and a combination of the two \cite{wang2007optimized,chen2013image}. In sampling-based approaches, a foreground-background sample pair is picked from few candidate samples taken from $F$ and $B$ regions by optimizing an objective function. This $(F,B)$ pair is then used to estimate $\alpha$ at a pixel with color $I$ by
\begin{equation}\label{eq:est}
\alpha =\frac{(I-B)\cdot(F-B)}{\left \| (F-B) \right \|^{2}},
\end{equation}
where $\left \| \cdot  \right \|^{2}$ denotes the Euclidean distance. $\alpha$-propagation based methods assume correlation between the neighboring pixels under some image statistics and use their affinities to propagate alpha values from known regions to unknown ones. The third category includes methods in which matting is cast as an optimization problem, where the color sampling component forms the data term and the alpha propagation component forms the smoothness term; solving for the alpha matte becomes an energy minimization task.
\begin{figure*}[t]
\centering
\includegraphics[width=1\linewidth, clip=true, trim=0.7cm 6.6cm 1.7cm 0.5cm]{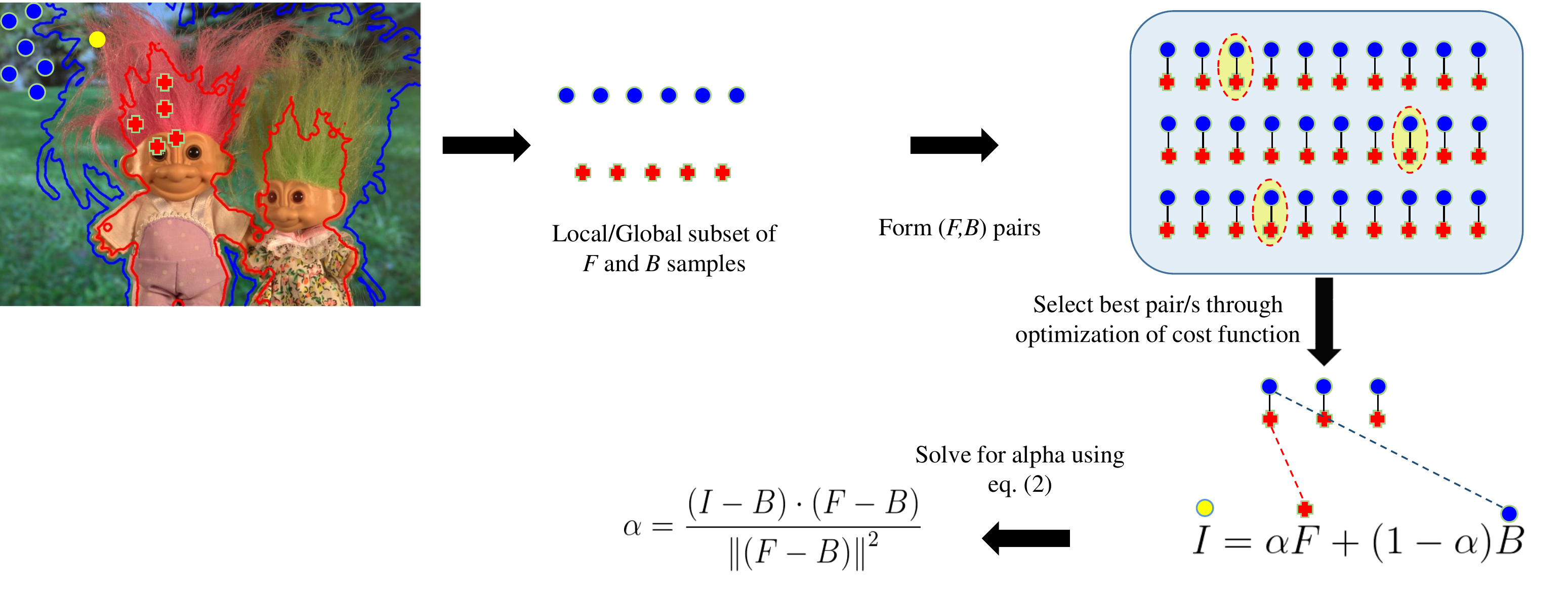}\\
(a) Existing sampling methods\\
\vspace*{0.5cm}
\includegraphics[width=1\linewidth, clip=true, trim=0.2cm 8.2cm 1.7cm 0.4cm]{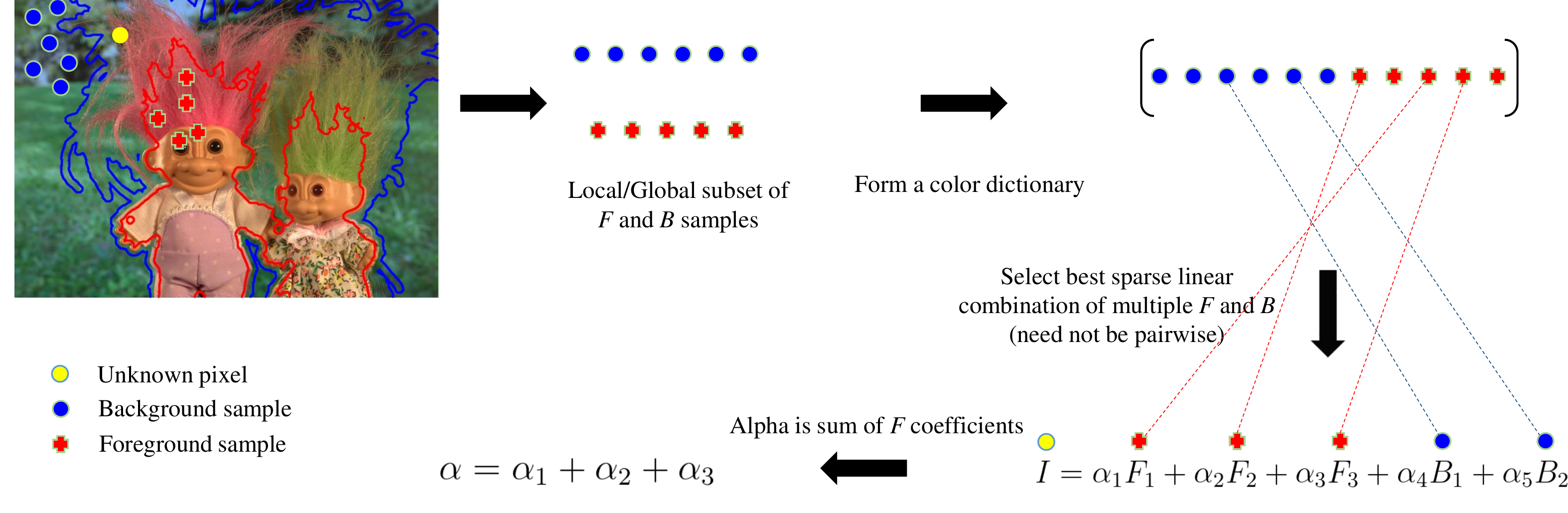}\\
(b) Proposed method
\caption{Illustrative comparison of matte estimation between methods using paired samples and the proposed approach. (a) Pairwise approaches use pair/s of $F$ and $B$ samples to estimate $\alpha$ at a pixel. (b) The proposed approach eliminates the need to restrict matte estimation using $(F,B)$ pairs. Instead, samples that produce least sparse reconstruction error at a pixel are selected.}
\label{fig:sampling_comparison}
\end{figure*}
The extension to videos is not straightforward as the human visual system is highly sensitive to jitter and flickering across frames. This phenomenon is observed when an image matting algorithm is applied to every frame of a video sequence. Hence, video matting also needs to address \emph{temporal coherence} for the mattes. However, video presents the opportunity to leverage information from surrounding frames in the estimation of the matte for a given frame.
  
The method proposed in this paper is based on sampling. However, there is a fundamental difference from other sampling-based approaches in the way alpha is estimated as illustrated in Fig.~\ref{fig:sampling_comparison}. For a given unknown pixel in yellow, all sampling methods collect a subset of labeled $F$ and $B$ samples in red and blue respectively, from a local/global region depending on the sampling strategy. Current methods form an exhaustive set of $(F,B)$ pairs by pairing each $F$ sample with all the other $B$ samples as illustrated in Fig.~\ref{fig:sampling_comparison}(a). The $(F,B)$ pair that best 'represents' the true foreground and background is chosen from candidate pairs through optimization of a cost function. $\alpha$ is then determined by eq.~(\ref{eq:est}). It is worth mentioning that Robust matting~\cite{wang2007optimized} selects 3 best pairs instead of just one and averages the alpha obtained for each pair. However, alpha is still estimated using $(F,B)$ pairs. In the proposed method, estimation of $\alpha$ is not governed by eq.~(\ref{eq:est}); instead, for a given unknown pixel, multiple $F$ and $B$ \emph{unpaired} samples that best reconstruct it, are picked from a subset of labeled samples. The sum of the reconstruction coefficients directly gives $\alpha$. This reinterpretation allows the matting framework to determine $\alpha$ based on more relevant $F$ and $B$ samples than with only one of each. The proposed framework is computationally faster as it removes the need of an expensive 2-D search for the best $(F,B)$ pair as in \cite{shahrian2013improving,he2011global,shahrian2012weighted}.    

\begin{figure}[t]
\includegraphics[width=0.95\linewidth, clip=true, trim=0.2cm 5.0cm 8.8cm 4.1cm]{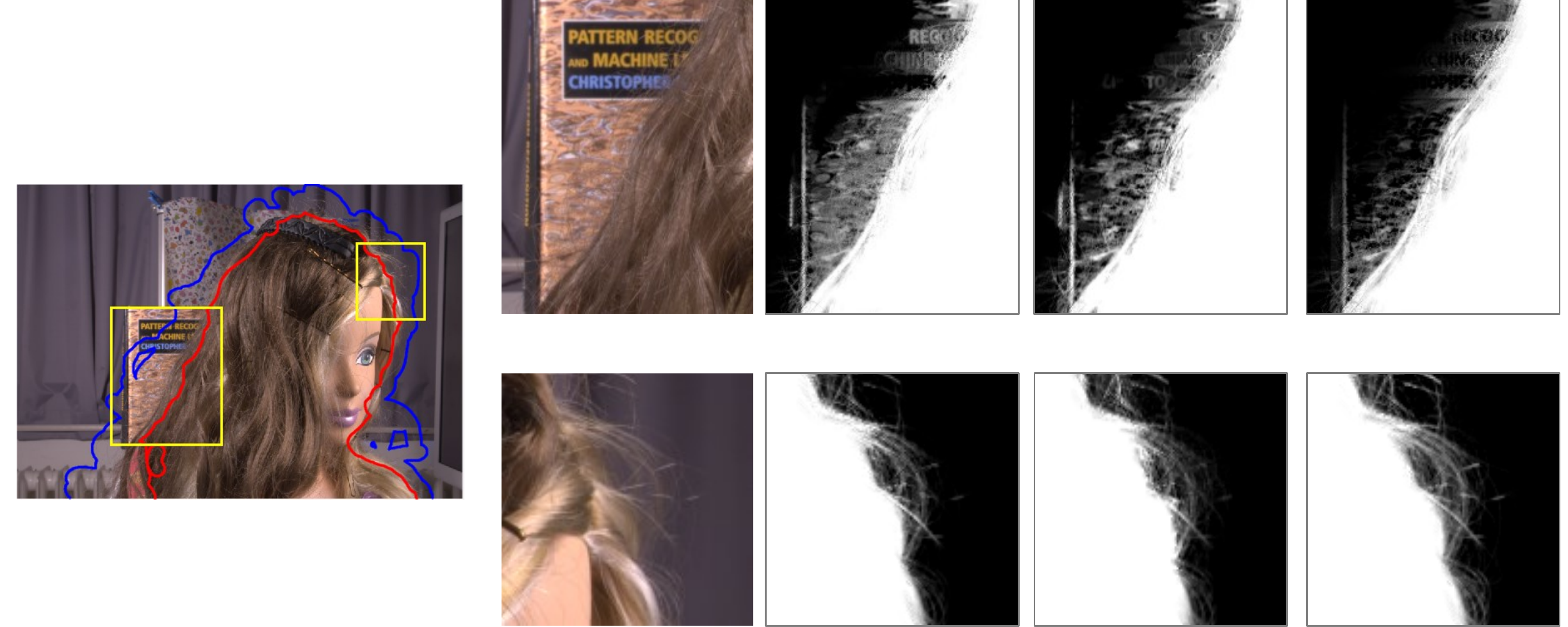}\\
\hspace*{1.15cm}(a)\hspace*{1.5cm} (b)\hspace*{0.9cm} (c)\hspace*{0.85cm} (d)\hspace*{0.99cm} (e)
\caption{Visual comparison of alpha matte generated from multiple unpaired $F$ and $B$ samples using sparse coding with other state-of-the-art sampling methods that employ ($F,B$) pairs. (a) Original image with foreground (red) and background (blue) boundaries, (b) zoomed regions, (c) Comprehensive sampling~\cite{shahrian2013improving}, (d) Weighted Color and Texture~\cite{shahrian2012weighted} and (e) proposed method.}
\label{fig:doll_comparison}
\end{figure}

The tool that enables the selection of multiple samples from a set to minimize the reconstruction error is feature coding.
 Many coding schemes have been proposed earlier such as, hard-assignment coding (HC)~\cite{lazebnik2006beyond}, localized soft-assignment coding (LSC)~\cite{liu2011defense}, sparse coding (SC)~\cite{wright2010sparse} and locality-constrained coding (LLC)~\cite{wang2010locality}. 
 The success of feature coding in various computer vision tasks such as face recognition~\cite{wright2009robust}, super-resolution~\cite{yang2010image} and image classification~\cite{yang2009linear} can be attributed to the fact that images generally lie on low-dimensional subspaces or manifolds from which representative samples could be picked based on an appropriate basis \cite{wright2010sparse}. In the proposed method, a dictionary of color values of $F$ and $B$ pixels is employed to determine the codes for a pixel in an unknown region. The sum of the normalized sparse codes for $F$ pixels directly provides the $\alpha$. Fig.~\ref{fig:doll_comparison}(a) shows an input image with 2 windows--one depicting highly textured and the other, smooth hairy regions. Pairwise methods are unable to pull an accurate matte due to the texture in the book cover as seen in the top row. The use of an explicit texture term in ~\cite{shahrian2012weighted} (Fig.~\ref{fig:doll_comparison}(d)) is insufficient to simultaneously handle the book pattern and the smoothness of the hair.  
The proposed method that makes use of multiple unpaired $F$ and $B$ samples (Fig.~\ref{fig:doll_comparison}(e)) is able to handle both the textured as well as the smooth parts to obtain a more accurate matte.
       
A preliminary version of this work was described in \cite{johnson2014sparse}. The major differences between this paper and [19] are (i) the sparse coded alpha estimate at a pixel is combined with its feature-space neighbors and spatial neighbors in a weighted graph model and solved in closed-form to obtain the final matte. This modification leads to improved performance and outperforms the previous work~\cite{johnson2014sparse} considerably on the benchmark dataset. 
(ii) A framework for extension to videos is also described that builds on the same graph-based formulation, across a block of frames to extract temporally coherent video mattes. Quantitative and qualitative  comparisons with other video matting approaches are provided. (iii) Finally, the applicability of other feature coding methods to the matting problem are presented and compared.  The key contributions are:\\
(i) To demonstrate feature coding as an alternative to the conventional compositing equation followed by sampling-based methods and\\
(ii) Temporally coherent matte extraction for videos by solving the matte for a block of frames together, using a weighted graph model.  
                  
The remainder of the paper is organized as follows. We review related work in section~\ref{sec:related}, followed by description of our image matting approach in section~\ref{sec:methodology}. The extension of our method to videos is detailed in section~\ref{sec:video_extension}. Experimental results are discussed in section~\ref{sec:results}, and we conclude the paper in section~\ref{sec:conclusion}.

\section{Related Work}
\label{sec:related}
Sampling-based matting methods can be divided into parametric and non-parametric methods. Parametric methods~\cite{chuang2001bayesian,wang2005iterative} describe the foreground and background samples as arising from parametric low-order statistical models and estimate alpha based on the distance of the unknown pixels to the known color distributions. They generate large fitting errors for textured regions where it is insufficient to model the higher-order statistics of color distribution. It tends to produce weak mattes when the trimap is coarse, leading to unreliable correlations between unknown pixels and known samples. Non-parametric methods~\cite{he2011global,gastal2010shared,shahrian2012weighted,shahrian2013improving} collect a subset of known $F$ and $B$ samples and estimate alpha from the best ($F,B$) pair, found through an optimization process. Different sampling strategies and final pair-selection criteria distinguish these methods. Samples are collected from spatially nearest boundary pixels~\cite{wang2007optimized}, by shooting rays from the unknown to the known pixels~\cite{gastal2010shared}, by selecting all the pixels on the known region boundaries~\cite{he2011global}, or by selecting a comprehensive set of samples from within the known regions through Gaussian mixture model (GMM) based clustering~\cite{shahrian2013improving}. The final $(F,B)$ pair, found through optimization of an objective function, controls the quality of the matte. Such approaches fail to produce a good matte when the $F$ and $B$ samples are nearby in the color space, or when the collected sample set fails to correlate with the actual color at the unknown pixel. Texture is used in addition to color as a feature for matting in \cite{shahrian2012weighted,varnousfaderani2013weighted} to address the problem of overlap in color distributions of $F$ and $B$.
Chen \textit{et~al}.~\cite{chen2013image} formulate the sampling-based alpha estimate in \cite{wang2007optimized}, smoothness Laplacian, along with locally linear embedding as a weighted graph and obtain a closed-form solution for the matte. Zhang \textit{et~al}.~\cite{zhang2012learning} treat matting as a supervised learning problem and use support vector regression to learn the alpha-feature model from known samples. This method is also affected by similar $F$ and $B$ colors since the learned feature model could be inaccurate. 

The only other method that uses sparsity for matting is compressive matting \cite{yoon2012alpha}. There are two important differences between their method and ours. First, they use a fixed window from which the dictionary is formed. This severely restricts the possibility of obtaining a very incoherent dictionary since images are generally smooth over a small neighborhood. Second, they map the sparse codes to the $\alpha$ values using a ratio of the $l_2$-norm of the sparse codes of $F$ pixels to the sum of $l_2$-norms of sparse codes of $F$ and $B$ regions. In our approach, by using special constraints on the sparse representation, the sum of the sparse codes for $F$ directly give $\alpha$. Finally, they do not present quantitative results on test images from \cite{alphawebsite} which contains a standard benchmark dataset on which state-of-the-art matting algorithms are evaluated. 

$\alpha$-propagation relies on the affinity between neighboring pixels to propagate the matte. Levin \textit{et~al}.~\cite{levin2008closed} used a color line model in a small neighborhood of pixels to propagate $\alpha$ across  unknown regions. The color line model assumption does not hold in highly textured regions due to strong edges that block the propagation of alpha. KNN matting~\cite{chen2012knn} considers nonlocal principle to formulate the affinities among K-nearest neighbors in a nonlocal neighborhood. Similar strategies to construct the Laplacian are employed in \cite{beiiterative2013,shi2013color} to overcome the limitation of the color line assumption. However, the smoothness assumption is insufficient to deal with complex images as high correlation among similar $F$ and $B$ colors wrongly propagates alpha. An extensive survey on image matting is available in~\cite{wang2008image}.

Conventional video matting systems~\cite{chuang2002video,wang2005interactive,li2005video,bai2009video} follow a two-step approach towards extracting the matte. First, a trimap is generated for each frame through user interaction. Many methods~\cite{wang2005interactive,li2005video,bai2009video} perform binary segmentation using graph cut, after which, morphological operations near the boundary generate a trimap. Optical flow is used in some approaches~\cite{chuang2002video,bai2011towards} to propagate the trimap to intermediate frames. Once the trimap is obtained for each frame, image matting algorithms are then applied to extract the matte.
 However, simply applying single-frame methods will not enforce temporal coherence by ignoring information contained in nearby frames.

$\alpha$-propagation methods have been extended to videos to maintain temporal coherence. Bai \textit{et~al}.~\cite{bai2009video} use robust matting~\cite{wang2007optimized} at each frame, followed by altering the matting Laplacian to bias towards alpha from the previous frame. Filtering the alpha matte temporally using level-set based interpolation is proposed in \cite{bai2011towards} to ensure temporal coherence. The disadvantage of such post-processing is that it reduces the spatial accuracy of the matte by smoothing the edges. Similar to images, $\alpha$-propagation approaches fail if the neighborhood for a pixel is too distant in space or time. \cite{lee2010temporally} extends \cite{wang2007optimized} to video by sampling from the previous and next frames. Ehsan \textit{et~al}.~\cite{sharianvideo} extended \cite{shahrian2013improving} to videos by selecting the best $(F,B)$ pair using an objective function that enforces temporal coherency in the sample set. However, the alpha at a pixel is still solved independently, and a post-processing step is used to maintain neighboring consistency. Motion information is used to modify the KNN Laplacian~\cite{chen2012knn} by considering two consecutive frames in building the affinity matrix for matting in \cite{li2013motion}.           
\begin{figure}[t]
\centering
\includegraphics[width=0.98\linewidth, clip=true, trim=0.2cm 11.05cm 0.2cm 2.4cm]{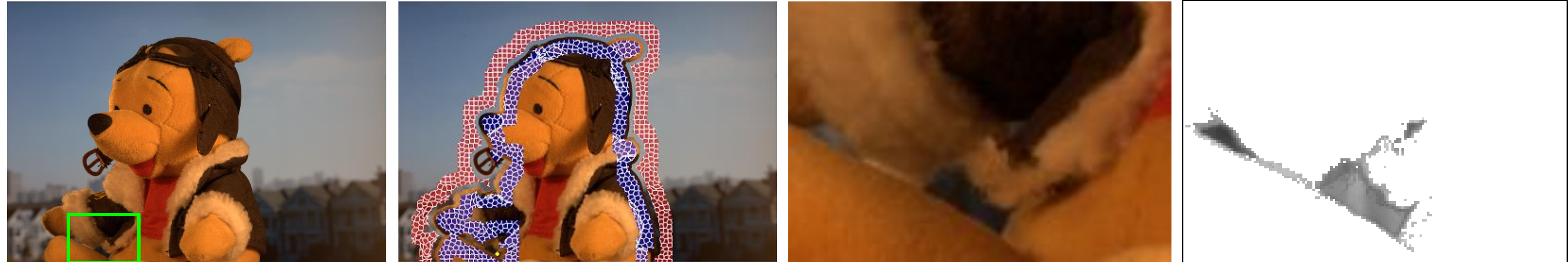}\\
(a)\hspace*{1.6cm} (b) \hspace*{1.6cm}(c) \hspace*{1.6cm} (d)
\caption{Illustration of sampling strategy and foreground probability map. (a) Input image, (b) superpixel samples, (c) zoomed window and its (d) foreground probability map.}
\label{fig:posterior}
\end{figure}
\section{Proposed method}
\label{sec:methodology}
As indicated earlier, the motivation of the proposed method is to determine $\alpha$ from a bunch of $F$ and $B$ samples in a sparse coding framework so that the codes directly provide the $\alpha$ values. The unknown pixels are categorized into high-certainty and low-certainty to allow for a comprehensive set of samples for dictionary formation. This is done through probabilistic segmentation of the image. Sparse coding at each pixel using the appropriate dictionary yields the $\alpha$ value. 

We use a pre-processing step to expand the known regions to unknown regions based on certain chromatic and spatial thresholds. An unknown pixel $i$ is considered as foreground if, for a pixel $j\in F$~\cite{shahrian2013improving}
\begin{equation}
(D(i,j)<E_{thr})\wedge (\left \| I_{i}-I_{j} \right \|\leq (C_{thr}-D(i,j)),
\end{equation}
where $D(i,j)$ is the Euclidean distance between pixel $i$ and $j$ in the spatial domain, $I_i$ is the color at pixel $i$, and $E_{thr}$ and $C_{thr}$ are thresholds empirically set to 12 and 4, respectively. A similar formulation is applied to expand background regions.

\subsection{Certainty of unknown pixels}
\label{subsec:conf}
We wish to classify an unknown pixel as high-certainty or low-certainty depending on how well the colors of foreground and background regions are separated in the neighborhood of the unknown pixel. If an unknown pixel is labeled high-certainty, then it implies that the $F$ and $B$ colors in its neighborhood are well separated and consequently, the size of the dictionary can be small since we can then ensure that the dictionary will be formed from highly incoherent samples. Overlapping of $F$ and $B$ color distributions is one of the problems that recent sampling based approaches like  \cite{shahrian2013improving} try to address. Pixels with low-certainty come from areas that have potentially complex and overlapping color distributions requiring a larger dictionary so that the variability in color can be captured.

The feature vector used for coding is the 6-D vector $[\:R\; G \;B \;L\; a \;b\:]^{T}$ consisting of the concatenation of the two color spaces RGB and CIELAB respectively. In order to reduce the sample space, we cluster labeled $F$ and $B$ pixels in a band of width 40 pixels along the boundary of the unknown region into superpixels using SLIC algorithm~\cite{achanta2012slic}.  The mean vector of each superpixel represents the $F$ and $B$ samples that make up the universal set. Fig.~\ref{fig:posterior}(a) shows an original image and Fig.~\ref{fig:posterior}(b) shows the universal set of superpixels from the foreground and background regions in blue and red respectively. From the universal set, samples are chosen to form dictionaries based on the certainty of the pixel.

Since the feature used for coding is color and the complexity of a region for matting is dependent on the overlap of foreground and background colors, we use probabilistic segmentation as a cue to label the unknown pixel as having low/high-certainty.
We adopt a non-parametric sampling-based probability measure~\cite{ju2013progressive} to 
determine the probability of a given pixel $i$ belonging to the foreground as
\begin{equation}
p(i)=\frac{p_{f}(i)}{p_{f}(i)+p_{b}(i)},
\label{eq:fgprob}
\end{equation}
where $p_{f}(i)$ is the foreground color affinity value given by
\begin{equation}
p_{f}(i)=exp\left (-\frac{\sum_{k=1}^{m}\left \| c(i)-c(f_{k}) \right \|^{2}}{m\cdot \delta }  \right ).
\end{equation}

Here $c(\cdot)$ is the RGB color value, $m$ is the number of spatially close foreground samples (closeness is measured in terms of Euclidean distance from unknown pixel to centroid of a superpixel) and $\delta$ is a weighting constant. In our experiments, we set $m$ and $\delta$ as 10 and 0.1 respectively. A higher value of $p_{f}(i)$ indicates that the pixel has higher affinity to $F$ samples. A similar expression is applicable to $p_{b}(i)$.

\begin{figure}[t]
{\centering
\includegraphics[width=0.98\linewidth, clip=true, trim=0.2cm 5cm 5.5cm 4.3cm]{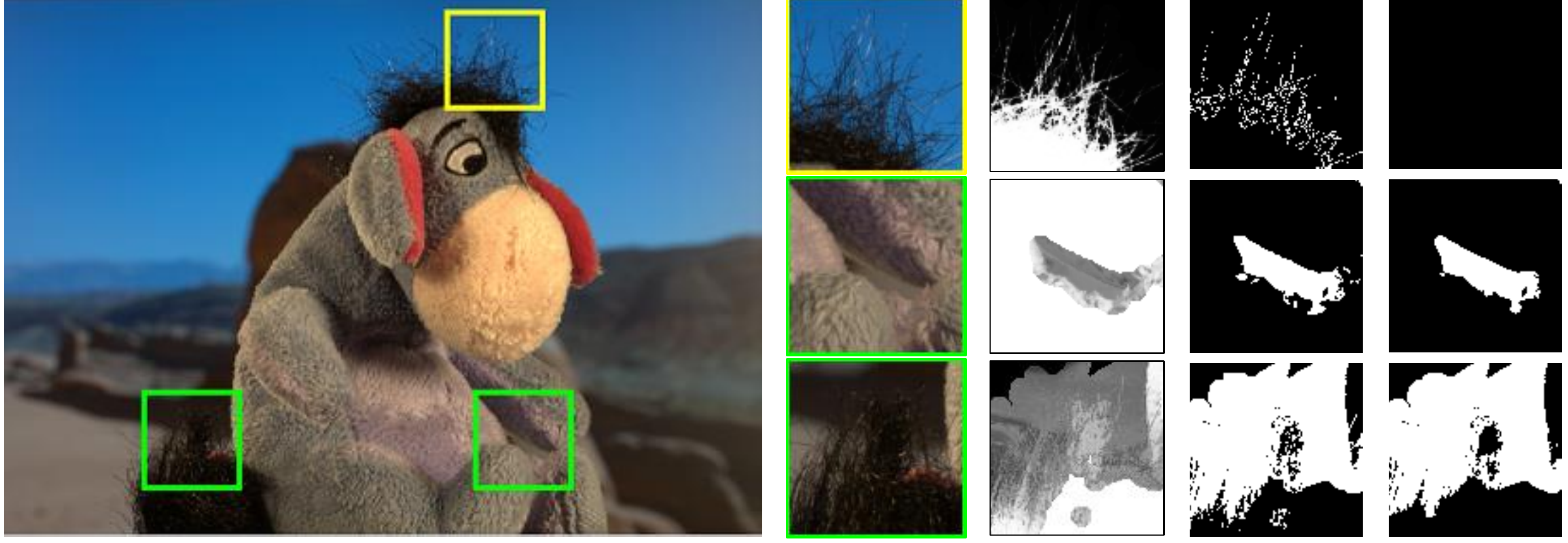}}\\
\hspace*{2cm}(a)\hspace*{2.14cm} (b) \hspace*{0.6cm}(c) \hspace*{0.5cm} (d)\hspace*{0.65cm}(e)
\caption{Defining low-certainty regions in complex areas. (a) Input image and (b) zoomed patches, corresponding (c) probability map, (d) thresholded mask for pixels in $[0.3,0.7]$ and (e) low-certainty mask for each region.}
\label{fig:lowconf}
\end{figure}

Eq.~(\ref{eq:fgprob}) provides a probability value on whether a pixel belongs to the foreground. Fig.~\ref{fig:posterior}(c) is a zoomed region of Fig.~\ref{fig:posterior}(a) where there is an overlap in the foreground and background color distributions. Fig.~\ref{fig:posterior}(d) shows the probability map with higher intensity denoting higher probability for a pixel to belong to foreground. In areas of complex color distributions, the distance in color space of the unknown pixel to foreground and background pixels are similar.  It has been observed that in such cases, the probability value as given by eq.~(\ref{eq:fgprob}) ranges from about 0.3 to 0.7. However, it is also observed that in regions with thin hairy structures whose color is significantly different from the background, the blending of colors causes the probability to also lie between 0.3 and 0.7. In order to distinguish such regions where the color distributions do not overlap, we consider a 7$\times$7 neighborhood and classify the unknown pixel as \emph{low-certainty} if the number of pixels with $p(i)$ in $[0.3,0.7]$ is larger than 35 (fixed empirically).   Fig.~\ref{fig:lowconf}(a) shows an image marked with regions having foreground probability in $[0.3,0.7]$. Fig.~\ref{fig:lowconf}(b) shows the marked windows separately. The green windows are areas of complex color distributions. However, the yellow window at the top shows a hairy region with separable $F$ and $B$ colors. The zoomed regions of the probability map in Fig.~\ref{fig:lowconf}(c) show that even in the hairy region, we can observe probabilities in $[0.3,0.7]$ as shown in the thresholded mask in Fig.~\ref{fig:lowconf}(d) . The additional neighborhood condition helps us to classify only the true complex color areas as the low-certainty mask, as shown in Fig.~\ref{fig:lowconf}(e) in which 1 is assigned to every low-certainty pixel and 0 to high-certainty pixels.

\subsection{Sparse coding to generate $\alpha$ }
\label{subsec:dictionary}
The size of the dictionary for an unknown pixel depends on whether it is classified as \emph{low-certainty} or \emph{high-certainty}. If the pixel is of high-certainty, i.e., the color distributions of nearby $F$ and $B$ are well separated as determined by the probabilistic segmentation, then the dictionary is formed from 40 (chosen empirically) spatially closest $F$ and $B$ samples. Note that the samples here are the mean vectors of the spatially closest superpixels.  As mentioned earlier, in such regions, the samples would be sufficiently incoherent for sparse coding.

\begin{figure}[t]
{\centering
\includegraphics[width=1\linewidth, clip=true, trim=0.1cm 12.5cm 0.1cm 2cm]{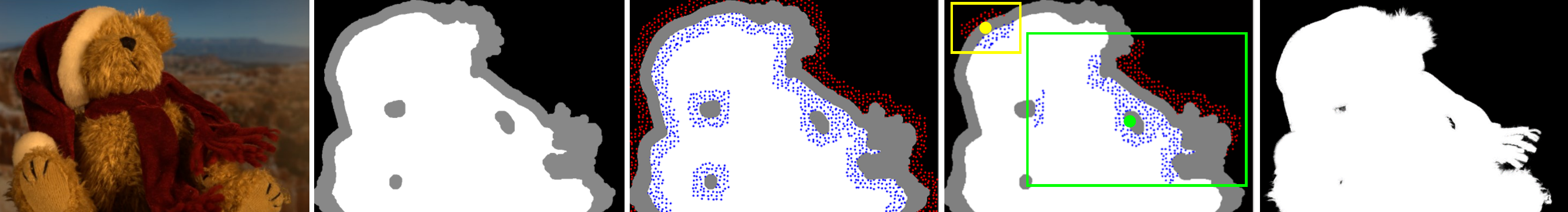}}\\
\hspace*{0.69cm}(a)\hspace*{1.23cm} (b) \hspace*{1.23cm}(c) \hspace*{1.13cm} (d)\hspace*{1.4cm}(e)
\caption{Dictionary formation for sparse coding to generate the matte. Foreground and background regions are shown in blue and red respectively. (a) Input image, (b) trimap, (c) universal set of samples, (d) dictionary for a low-certainty pixel (green window) and high-certainty pixel (yellow window) and (e) estimated alpha.}
\label{fig:refinement}
\end{figure}

The low-certainty pixels are potential areas of overlapping color distributions. In this case, the dictionary size is larger. Thus, for a given unknown pixel, one-third of the superpixels from the universal set that are closest in terms of Euclidean distance in the spatial domain constitute the dictionary. Fig.~\ref{fig:refinement}(a) and Fig.~\ref{fig:refinement}(b) show an input image and its corresponding trimap. The universal sample set of $F$ and $B$ regions in blue and red respectively, is shown in Fig.~\ref{fig:refinement}(c). For a given unknown pixel of low-certainty shown in green in Fig.~\ref{fig:refinement}(d), the dictionary is a larger subset of the universal sample set than that of a high-certainty pixel in yellow.

The 6-D color vector, which forms the feature vector for coding, is normalized to unit length. The dictionary is formed by concatenating the $F$ and $B$ samples horizontally as $\mathbf{D}=[F_1,F_2,...F_n ,B_1,B_2,..B_n]$ which enables us to determine the location of each $F$ and $B$ atom during sparse coding. Given the dictionary $ \mathbf{D} $ for an unknown pixel $i$, its sparse code is determined as
\begin{equation}\label{eq:spsmatte}
\begin{split}
\beta=argmin\left \| v_{i}-\mathbf{D}\beta_{i} \right \|_{2}^{2} \;\;\;\;\;\;   s.t\;\;  \left \| \beta_{i} \right \|_{1}\leq 1 \;;\; \beta_{i}\geq 0,  
\end{split}
\end{equation}
where $v_{i}$ is the feature vector at $i$. The sparse codes $\beta_i$  are generated using a modified version of the Lasso algorithm~\cite{mairal2010online}. The sparse coding procedure is presented with an appropriate set of $F$ and $B$ samples and the sparse coefficients sum up to less than or equal to 1. It has been observed that for most of the pixels, the sparse codes add up to 1. For the few exceptions, the sparse coefficients are normalized to sum up to 1. For example, in GT04 image, out of the 124604 unknown pixels that underwent sparse coding, only 44 pixels had to be normalized as the corresponding codes did not sum up exactly to 1. In order to avoid negative values, the second constraint forces all coefficients to be positive. The sparse codes corresponding to atoms in the dictionary that belong to foreground regions are added to form the alpha for the unknown pixel i.e.
\begin{equation}
\hat{\alpha}=\sum_{p \in F} \beta^{(p)}.
\end{equation}
 Hence, the sparse codes directly provide the value of alpha. Fig.~\ref{fig:refinement}(e) shows the alpha matte extracted from the sparse codes using our approach on the input image in Fig.~\ref{fig:refinement}(a). A good quality matte is obtained even when the unknown region is well inside the foreground which is a challenge for propagation-based methods. 
\subsection{Graph model-based matte optimization}
\begin{figure}[t]
\centering
\includegraphics[scale=0.4, clip=true, trim=9.1cm 2.5cm 10.9cm 1.7cm]{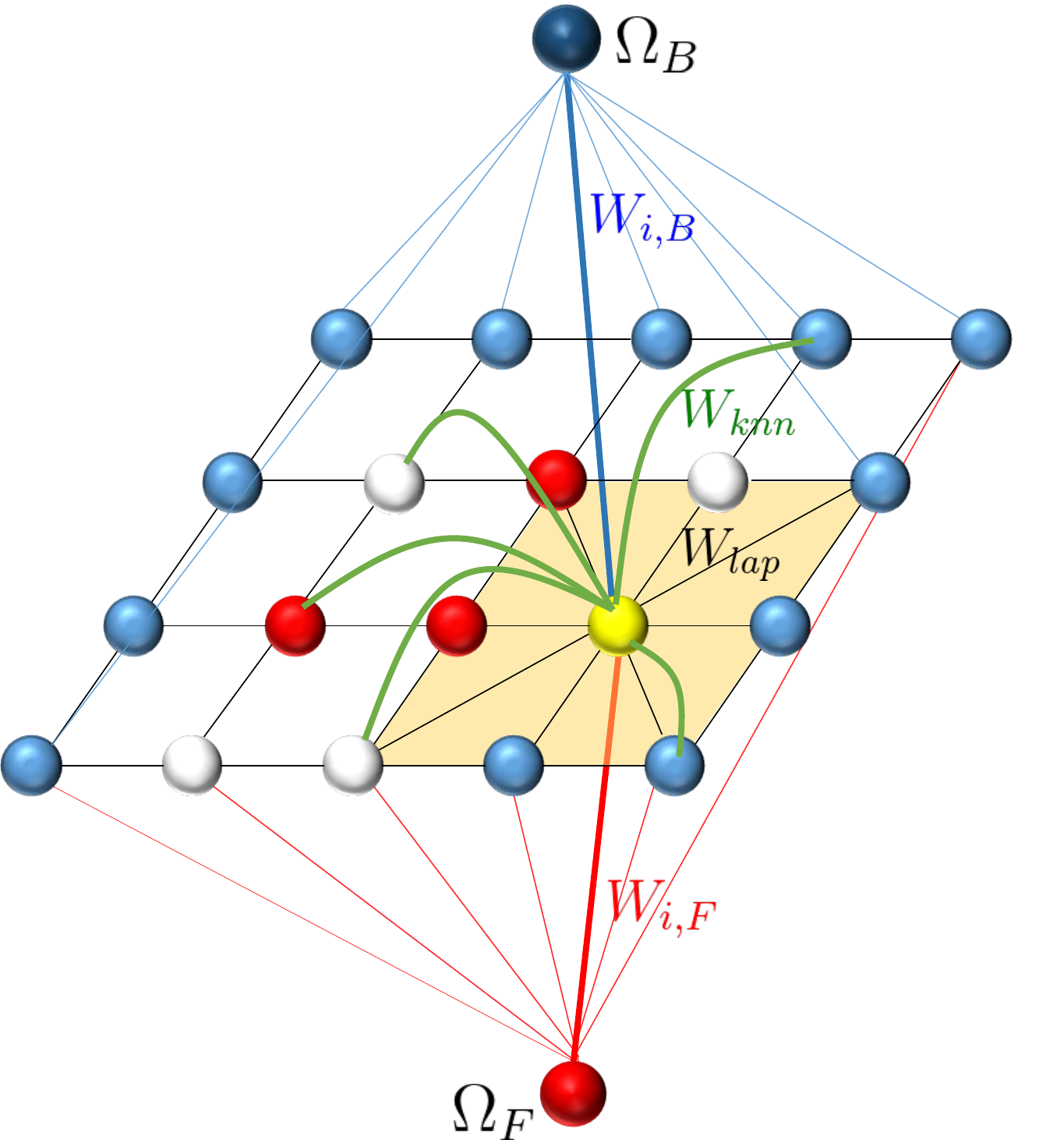}\\
\caption{Illustration of graph model on image for optimization of the matte. Each pixel is connected to its spatial neighbors ($W_{lap}$), feature-space neighbors ($W_{knn}$) and to the virtual foreground ($W_F$) and background ($W_B$) nodes.}
\label{fig:image_graph}
\end{figure}

As with other sampling-based approaches, the correlation between the neighboring pixels are ignored while estimating the matte at each pixel. Post-processing is usually employed to smooth out the matte as in \cite{shahrian2012weighted,shahrian2013improving}. We use the  graph-based optimization followed by \cite{wang2007optimized,chen2013image} to obtain our final matte.  
 A smoothness term consisting of the matting Laplacian~\cite{levin2008closed} and the $K$-nearest neighbors~\cite{chen2012knn} is combined with the initial sparse coded estimate $\hat{\alpha}$ as the data term in a graph model, and solved in closed-form as a sparse system of linear equations. Fig.~\ref{fig:image_graph} illustrates our graph model where red, blue and white nodes represent the pixels marked by the trimap as foreground, background and unknown respectively. Two virtual nodes $\Omega_{F}$ and $\Omega _{B}$, representing the foreground and background are connected to each pixel through the data weights $W_{i,F}$ and $W_{i,B}$ respectively.  

We associate the initial estimate of the matte with a confidence value $\gamma$, which together indicates whether a given pixel belongs to the foreground or background.  The data weights for pixel $i$ are defined as
 \begin{equation}
 W_{i,F}=\gamma \hat{\alpha_i },\qquad \qquad \qquad
 W_{i,B}=\gamma (1-\hat{\alpha_i }),
 \end{equation}
 where a true foreground pixel should have higher value of $W_{i,F}$ and lower value of $W_{i,B}$, and vice versa.
 As $\gamma$ encodes the confidence in the initial estimate, a larger weight is assigned to more confident estimates.
The confidence value at a pixel $i$ is 
\begin{equation}
\gamma(i)=\gamma_{sprec}(i)\cdot \gamma_{colrec}(i),
\end{equation}
where $\gamma_{sprec}(i)= e^{-\left \| v_{i}-\hat{v}_{i} \right \|^{2}}$ measures the confidence in reconstructing the input feature vector based on the sparse coefficients ($\hat{v}_i=\mathbf{D}\beta_i$), and $
\gamma_{colrec}(i)= e^{-\left \| I_{i}-[F_{i}^{rgb}\;B^{rgb}_{i}]\hat{\mathbf{\alpha}}_{i}  \right \|^{2}}
$ measures the chromatic distortion.

To enforce smoothness constraint across the image, each pixel is connected to its spatial neighbors in a $3\times3$ window through the weights $W_{lap}$, and to its $K$-nearest neighbors in the feature space through the weights $W_{knn}$. For pixels $i$ and $j$ in a $3\times3$ window $w_k$, the weight $W_{lap}$ is defined as~\cite{levin2008closed}

\begin{equation}
\begin{split}
W_{lap}(i,j)=\sum_{k\mid(i,j)\in w_k}  ( \delta_{ij}-\frac{1}{9}(1+(I_i-\mu_k)\\
(\Sigma_k+\frac{\epsilon }{9}I_3)^{-1}(I_j-\mu_k))  ),
\end{split}
\end{equation}
where $\delta_{ij}$ is the Kronecker delta, $\epsilon$ is a small regularization constant ($10^{-7}$), $\mu_k$ and $\Sigma_k$ are the mean and variance of colors in the window and $I_3$ is the $3\times3$ identity matrix. 
$K-$nearest neighbors at a pixel $i$ in the $RGBxy$ feature space are used to compute the affinity as~\cite{chen2012knn}   
 \begin{equation}
 W_{knn}(i,j)=1-\frac{\left \| X_i-X_j \right \|}{\sigma },
 \end{equation}      
where $j$ is a neighbor of $i$, and $\sigma$ is a weighting constant to constrain the values of $W_{knn(i,j)}$ in $[0,1]$. $K$ is fixed as 12 in our experiments. 

\begin{figure}[t]
\centering
\includegraphics[width=0.98\linewidth, clip=true, trim=0.1cm 3.1cm 0.2cm 3.5cm]{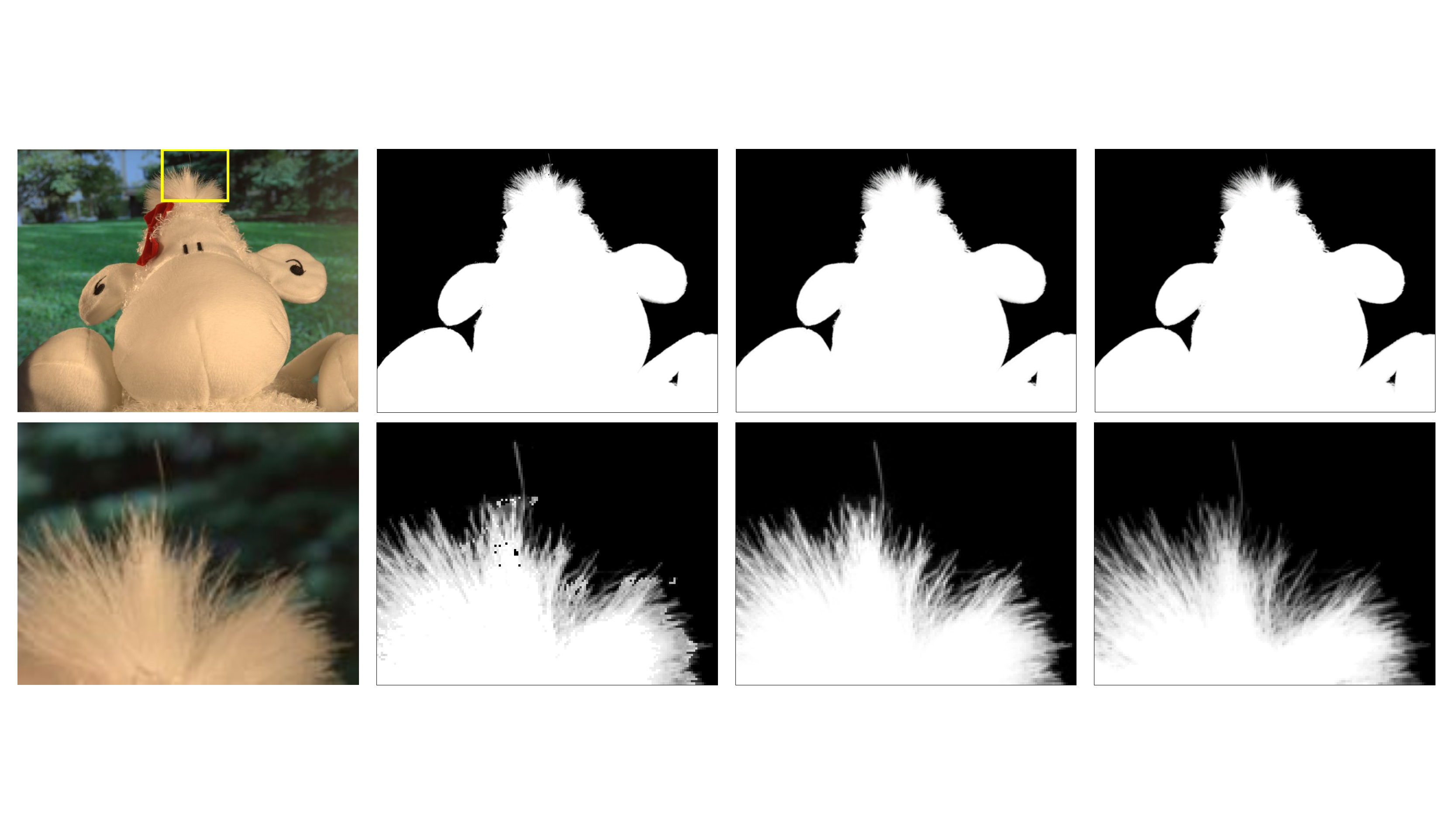}\\
(a)\hspace*{1.6cm} (b) \hspace*{1.6cm}(c) \hspace*{1.6cm} (d)
\caption{Visual effect of alpha refinement using the graph model. (a) Input image, (b) initial estimate, (c) final alpha matte and (d) ground truth.}
\label{fig:alpha_refinement}
\end{figure}

The energy function used for solving the matte is~\cite{chen2013image}
\begin{equation}
\begin{split}
E=\lambda \sum_{i\in \mathcal{V}}(\alpha_i-h_i)^2\; +  \sum_{i=1}^{N}\left (\sum_{j\in \mathcal{N}_i}W_{ij}(\alpha_i-\alpha_j )^2 \right ),
\end{split}
\end{equation}             
where $N$ is the total number of nodes (pixels) in the graph model and $\mathcal{V}$ is the set of definite foreground and background pixels. The first term ensures that the final matte is consistent with the user specified constraints while the second term ensures that neighboring pixels share similar alpha values. $h_i$ enforces the user defined constraints -- $0$ if $i$ is marked as definite background and $1$ if marked as foreground. $\lambda$ is fixed empirically to be 100. The set $\mathcal{N}_i$ represents the spatial and feature neighbors of a pixel $i$, along with the two virtual nodes. Hence, the weights $W_{ij}$ combine both the smoothness weights $W_{knn}$, $W_{lap}$ and the data weights $W_{i,F}$ and $W_{i,B}$ additively at each pixel, i.e., $W_{ij}=W_{knn}(i,j) + W_{lap}(i,j)+W_{i,F}+W_{i,B}$. The energy function is written in matrix form as
\begin{equation}
E=\lambda(\alpha -H)^T\Gamma (\alpha -H) + \alpha^TL^TL\alpha,
\label{eq:linsys}
\end{equation}
where 
\begin{equation*}
    L_{ij}= 
\begin{cases}
    W_{ii},\quad& \text{if } i=j,\\
    -W_{ij},\quad & \text{if}\; i\; \text{and}\; j\; \text{are neighbors},\\
    0,\quad              & \text{otherwise}.
\end{cases}
\end{equation*}
The weight $W_{ii}= \sum_{j\in \mathcal{N}_i} W_{ij}$, $\Gamma$ is a $N\times N$ diagonal matrix with $\Gamma_{ii}=1$ if $i\in \mathcal{V}$, else 0. $H$ is a $N\times 1$ vector with user constrained values.
Eq. (\ref{eq:linsys}) is a quadratic function in $\alpha$ and can be solved in closed-form as a sparse linear system.
 \begin{equation}
 \alpha=(L^TL+\lambda\Gamma)^{-1}H.
 \label{eq:solvelin}
 \end{equation}
 Fig.~\ref{fig:alpha_refinement} demonstrates the effect of our graph-based optimization in refining the initial alpha estimate. The irregularities formed in estimating alpha independently at each pixel (Fig.~\ref{fig:alpha_refinement}(b)) is eliminated, leading to high quality mattes visually similar to the ground truth.
\begin{figure}[t]
 	\begin{minipage}[b]{1\linewidth}
 		\centering
 		\includegraphics[width=0.99\linewidth,clip=true, trim=2.2cm 2.88cm 4.9cm 3.7cm]{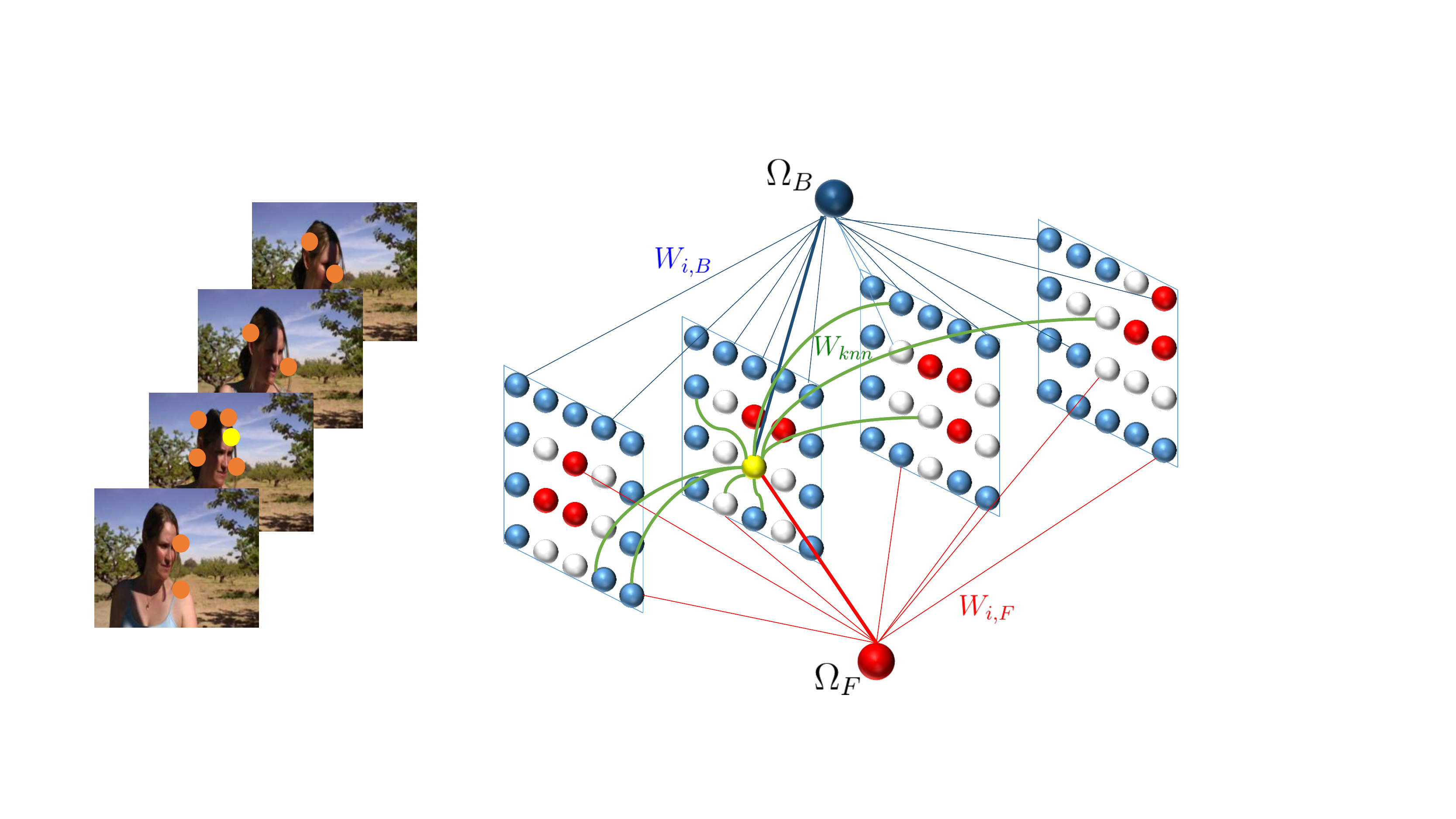}
 		%  \vspace{1.5cm}
 	\end{minipage}
 	\hspace*{1cm} (a) \hspace*{4cm} (b)
 	\caption{Graph model for temporally coherent video matting. (a) $K$-nearest neighbors of the pixel (yellow) in the second frame are encoded across multiple frames, (b) Each pixel (node) is connected to its feature space neighbors as well as the virtual foreground and background nodes in the graph.}
 	\label{fig:overview}
 \end{figure}
 \begin{figure*}[t]
 {\centering
 \includegraphics[width=0.99\linewidth, clip=true, trim=0.95cm 4.2cm 0.1cm 0.2cm]{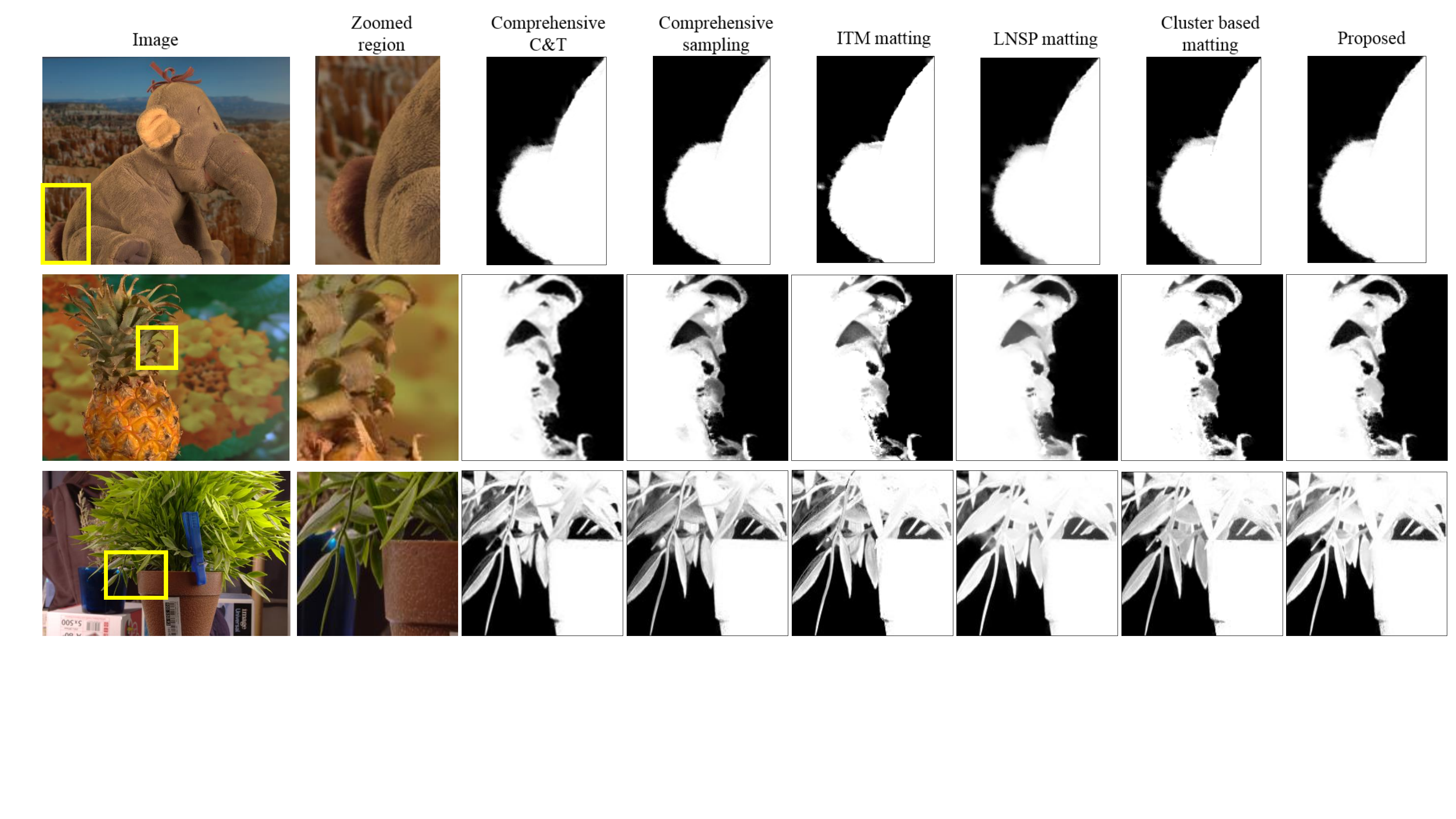}}
 \hspace*{1.4cm}(a)\hspace*{2.25cm}(b)\hspace*{1.75cm}(c)\hspace*{1.75cm}(d)\hspace*{1.65cm}(e)\hspace*{1.75cm}(f)\hspace*{1.65cm}(g)\hspace*{1.75cm}(h)
 \caption[]{Qualitative comparison of proposed method on elephant, pineapple and plant images with top 5 methods at~\cite{alphawebsite}. 
 (a) Input image, (b) zoomed windows, (c) Comprehensive weighted color and texture~\cite{varnousfaderani2013weighted}, (d) Comprehensive sampling~\cite{shahrian2013improving}, (e) Iterative transductive matting~\cite{beiiterative2013}, (f) LNSP matting~\cite{chen2013image}, (g) Cluster based matting\protect\footnotemark[1] and (h) proposed method. Zoomed in regions show the effectiveness of our method.}
 \label{fig:test_qualitative}
 \end{figure*}
\section{Extension to video}
\label{sec:video_extension}
The proposed sparse coding framework for matting is extended to videos to extract temporally coherent mattes by constructing the dictionary and graph model across $M$ consecutive frames. To achieve temporal coherence, the alpha at a pixel needs to be consistent not just with neighboring pixels in the same frame, but also across similar pixels from nearby frames~\cite{sharianvideo,bai2011towards}. Fig.~\ref{fig:overview} illustrates our multi-frame graph model for videos. Similar to the image matting framework, the white nodes are the unknown pixels for which the alpha is to be estimated. However, each pixel is connected to its $K$-nearest neighbors in the feature space across frames for ensuring temporal smoothness  as shown by the green lines in Fig.~\ref{fig:overview}(b). It is to be noted that we do not use spatial Laplacian smoothness term in our video graph formulation. As mentioned in section~\ref{sec:related}, the challenge in extending the matting Laplacian to the 3-D space arises from the oversmoothness caused by utilizing the spatio-temporal neighbors in defining the affinity. However, feature-space neighbors do not violate the assumption that similar features should share similar alpha values.
\footnotetext[1]{This method has not been published at the time of submission.}

For a given pixel $i$ at frame $t$, the dictionary for sparse coding is formed by sampling across the multi-frame block. The number of samples chosen from the current frame is the highest, with the number decreasing with increasing temporal distance. By utilizing the samples from nearby frames, a comprehensive sample set is ensured for the initial matte estimate. The confidence value for the initial estimate is identical to that used for image matting. The initial estimate and its confidence value forms the data term for the video model. To enforce temporal coherency across frames, each pixel is connected to its $K$-nearest neighbors in the feature space through the weights $W_{knn}$. Fig.~\ref{fig:overview}(a) illustrates the feature neighbors across 4 frames for the pixel marked yellow. We maintain the simple feature used for the image algorithm and do not utilize motion flow which is time consuming. The same energy function as eq.~(\ref{eq:linsys}) is used to determine the video matte across $M$ frames, except that now the affinity matrix $L$ is sparse and symmetric and of size $(M\cdot n+2)\times(M\cdot n+2)$, where $n$ is the number of pixels in a frame. We fix $M$ as 4 in our experiments. A higher number of frames can be used to form the multi-frame block, but the memory requirement is prohibitive due to the large size of the affinity matrix. The resulting cost function can be efficiently solved as a sparse linear system of equations using eq.~(\ref{eq:solvelin}) in the order of a few seconds for each frame. It is to be noted that the mattes of $M$ frames are solved simultaneously using our approach.       
%-------------------------------------------------------------------------

\section{Experimental Results}
\label{sec:results}
The effectiveness of the proposed method is evaluated using the benchmark dataset~\cite{rhemann2009perceptually} for image matting. It consists of 35 images covering a wide range of transparency of pixels - from opaque to fully transparent.  27 images form the training set with publicly available ground truth. The remaining 8 images form the test set whose ground truth is hidden from the public and is used for benchmark evaluation and ranking~\cite{alphawebsite}. The effectiveness of the proposed video matting method is evaluated on video sequences from the dataset introduced in \cite{sharianvideo}. Trimaps are generated on each frame of the sequence using \cite{bai2011towards}. 
\begin{figure}[t]
{\centering
\includegraphics[width=0.99\linewidth,clip=true, trim=2.2cm 2.1cm 5.5cm 1.5cm]{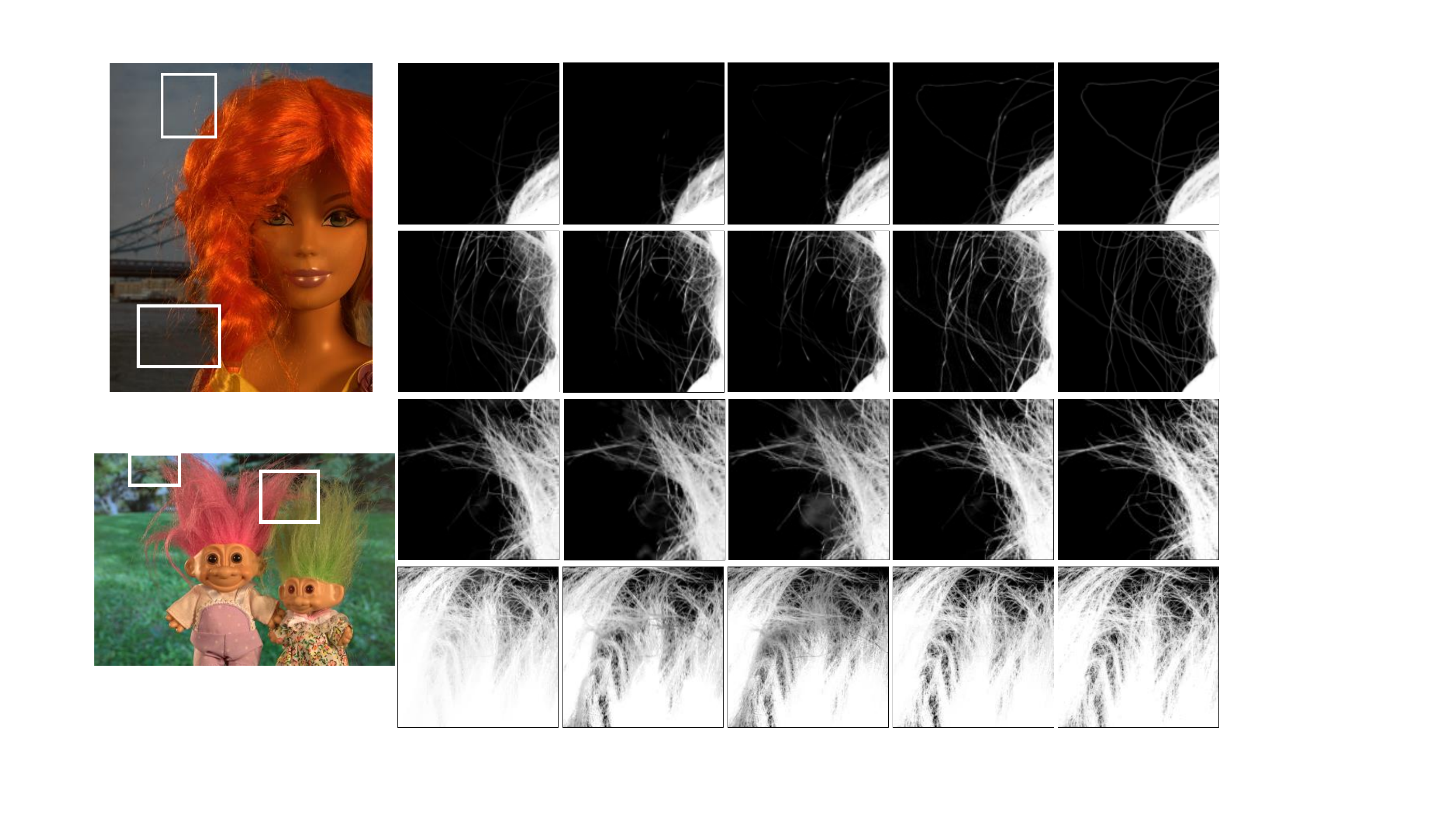}}
\hspace*{1cm}(a)\hspace*{1.4cm}(b)\hspace*{0.9cm}(c)\hspace*{0.9cm}(d)\hspace*{0.9cm}(e)\hspace*{0.9cm}(f)
 \caption{Qualitative comparison of proposed method against other state-of-the-art methods on the training dataset. (a) Image, zoomed regions showing the mattes of (b) Closed form~\cite{levin2008closed}, (c) Weighted color and texture~\cite{shahrian2012weighted}, (d) Comprehensive sampling~\cite{shahrian2013improving}, (e) proposed method and (f) ground truth.}
 \label{fig:train_qualitative}
 \end{figure}
\subsection{Image matting evaluation}
\subsubsection{Qualitative evaluation}
\begin{figure}[h]
{\centering
\includegraphics[width=1\linewidth, clip=true, trim=0.1cm 0.3cm 0.15cm 0.75cm]{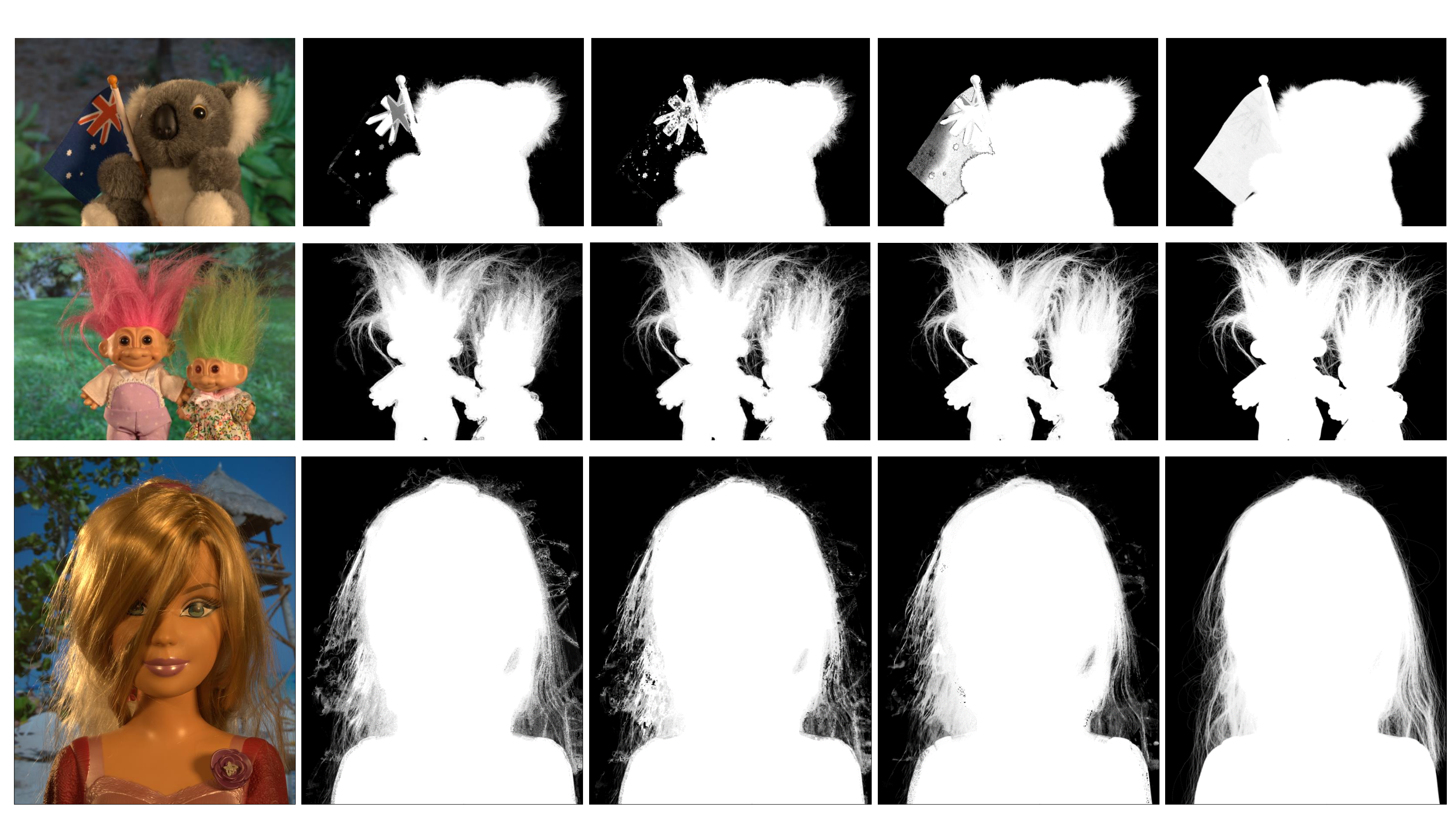}}
\hspace*{0.8cm}(a)\hspace*{1.35cm}(b)\hspace*{1.35cm}(c)\hspace*{1.35cm}(d)\hspace*{1.35cm}(e)
\caption{Qualitative comparison of recent sampling methods without pre and post-processing steps. (a) Input image, corresponding mattes of (b) Comprehensive sampling~\cite{shahrian2013improving} and (c) Weighted Color and Texture~\cite{varnousfaderani2013weighted} have more artifacts than (d) proposed method when compared to (e) ground truth.}
\label{fig:prepost}
\end{figure}

Fig.~\ref{fig:train_qualitative} shows the visual comparison of the proposed matting method with other recent state-of-the-art methods on two images from the training dataset. Closed-form matting~\cite{levin2008closed} oversmooths the matte, leading to loss of fine detail in hairy regions (Fig.~\ref{fig:train_qualitative}(b)), while  weighted color and texture matting~\cite{shahrian2012weighted} in Fig.~\ref{fig:train_qualitative}(c) uses texture feature in addition to color and is still unable to discriminate the finer structures in a complex background region. Comprehensive sampling~\cite{shahrian2013improving} (Fig.~\ref{fig:train_qualitative}(d)) utilizes a large sampling set, but results in smooth bands near the foreground-background boundary due to post-processing. The proposed method, shown in Fig.~\ref{fig:train_qualitative}(e), is able to extract a high quality matte by optimizing the sampling and smoothing terms which is visually closer to the ground truth given in Fig.~\ref{fig:train_qualitative}(f).

The test dataset contains the 8 most difficult subset of images from the dataset. Qualitative comparisons with state-of-the-art methods on the elephant, plant and pineapple images are shown in Fig.~\ref{fig:test_qualitative}. Additional results are presented in the supplementary material. The high overlap in $F$ and $B$ colors and complex texture make these images difficult for matting as shown by the zoomed patches in Fig.~\ref{fig:test_qualitative}(b). The presence of another plant in the background causes the texture to misconstrue the foreground as background in \cite{varnousfaderani2013weighted,shahrian2013improving} as seen in Fig.~\ref{fig:test_qualitative}(c,d) of the plant image. ITM matting~\cite{beiiterative2013} wrongly propagates the background into the leaves, the head of the pineapple, and the tail of the elephant in Fig.~\ref{fig:test_qualitative}(e). Comprehensive sampling~\cite{shahrian2013improving}, which uses a larger sampling set, is able to differentiate the elephant tail but still classifies the leaves and pineapple as mixed pixels in the last two rows of Fig.~\ref{fig:test_qualitative}(d). The smooth prior in LNSP matting~\cite{chen2013image} oversmooths the matte leading to a band between the foreground and background at the pineapple head in Fig.~\ref{fig:test_qualitative}(f). Our method is able to reduce such artifacts to extract a visually superior matte in these areas as shown in Fig.~\ref{fig:test_qualitative}(h).

\begin{table}[t!]
\caption{Ranks of different matting methods with respect to sum of absolute differences (SAD), mean squared error (MSE) and gradient error measures on benchmark dataset evaluated at \cite{alphawebsite}.}
\scriptsize
\begin{minipage}{0.5\textwidth}
\centering
\resizebox{\textwidth}{!}{%
\begin{tabular}{lcccc}
                                                              & \multicolumn{4}{c}{\textbf{SAD}}                                                                                                                                                                                                                                                                                                       \\ \hline
\multicolumn{1}{|l|}{Method}                                  & \multicolumn{1}{c|}{\begin{tabular}[c]{@{}c@{}}Avg. \\ small\\  rank\end{tabular}} & \multicolumn{1}{c|}{\begin{tabular}[c]{@{}c@{}}Avg.\\ large \\ rank\end{tabular}} & \multicolumn{1}{c|}{\begin{tabular}[c]{@{}c@{}}Avg.\\ user\\ rank\end{tabular}} & \multicolumn{1}{c|}{\begin{tabular}[c]{@{}c@{}}Overall\\ rank\end{tabular}} \\ \hline
\rowcolor[HTML]{CBCEFB} 
\multicolumn{1}{|l|}{\cellcolor[HTML]{CBCEFB}Proposed method} & \multicolumn{1}{c|}{\cellcolor[HTML]{CBCEFB}8.3}                                  & \multicolumn{1}{c|}{\cellcolor[HTML]{CBCEFB}8}                                  & \multicolumn{1}{c|}{\cellcolor[HTML]{CBCEFB}7.5}                                & \multicolumn{1}{c|}{\cellcolor[HTML]{CBCEFB}7.9}   
\\ \hline
\multicolumn{1}{|l|}{LNSP matting}                            & \multicolumn{1}{c|}{5.4}                                                             & \multicolumn{1}{c|}{7.1}                                                          & \multicolumn{1}{c|}{11.5}                                                       & \multicolumn{1}{c|}{8}                                                                             \\ \hline
\multicolumn{1}{|l|}{Comprehensive sampling}                                     & \multicolumn{1}{c|}{8.5}                                                          & \multicolumn{1}{c|}{9.5}                                                          & \multicolumn{1}{c|}{11.8}                                                        & \multicolumn{1}{c|}{9.9}                                                    \\ \hline
\multicolumn{1}{|l|}{Iterative Transductive matting}                  & \multicolumn{1}{c|}{11.4}                                                           & \multicolumn{1}{c|}{9}                                                          & \multicolumn{1}{c|}{9.8}                                                       & \multicolumn{1}{c|}{10}                                                    \\ \hline
\multicolumn{1}{|l|}{Comprehensive Weighted C\&T}                             & \multicolumn{1}{c|}{11}                                                          & \multicolumn{1}{c|}{10.5}                                                           & \multicolumn{1}{c|}{10}                                                        & \multicolumn{1}{c|}{10.5}                                                     \\ \hline
\multicolumn{1}{|l|}{SVR matting}             & \multicolumn{1}{c|}{13.1}                                                          & \multicolumn{1}{c|}{10.1}                                                          & \multicolumn{1}{c|}{9.3}                                                          & \multicolumn{1}{c|}{10.8}                                                   \\ \hline
\multicolumn{1}{|l|}{\cellcolor[HTML]{C0C0C0}Sparse coded matting}               & \multicolumn{1}{c|}{\cellcolor[HTML]{C0C0C0}13.5}                                                          & \multicolumn{1}{c|}{\cellcolor[HTML]{C0C0C0}11.3}                                                         & \multicolumn{1}{c|}{\cellcolor[HTML]{C0C0C0}8.1}                                                          & \multicolumn{1}{c|}{\cellcolor[HTML]{C0C0C0}11}                                                   \\ \hline
\multicolumn{1}{|l|}{Weighted Color \& Texture}                    & \multicolumn{1}{c|}{10.5}                                                           & \multicolumn{1}{c|}{13.6}                                                         & \multicolumn{1}{c|}{12.6}                                                       & \multicolumn{1}{c|}{12.3}                                                   \\ \hline
\multicolumn{1}{|l|}{CCM}                 & \multicolumn{1}{c|}{14.6}                                                            & \multicolumn{1}{c|}{11.6}                                                           & \multicolumn{1}{c|}{10.6}                                                       & \multicolumn{1}{c|}{12.3}                                                   \\ \hline
\multicolumn{1}{|l|}{Shared matting}          & \multicolumn{1}{c|}{12.6}                                                           & \multicolumn{1}{c|}{15.4}                                                          & \multicolumn{1}{c|}{11.9}                                                        & \multicolumn{1}{c|}{13.3}                                                    \\ \hline
\end{tabular}
}

\end{minipage}\\
\\
\\
\begin{minipage}{0.5\textwidth}
\centering
\resizebox{\textwidth}{!}{%
\begin{tabular}{lcccc}
                                                              & \multicolumn{4}{c}{\textbf{MSE}}                                                                                                                                                                                                                                                                                                       \\ \hline
\multicolumn{1}{|l|}{Method}                                  & \multicolumn{1}{c|}{\begin{tabular}[c]{@{}c@{}}Avg. \\ small\\  rank\end{tabular}} & \multicolumn{1}{c|}{\begin{tabular}[c]{@{}c@{}}Avg.\\ large \\ rank\end{tabular}} & \multicolumn{1}{c|}{\begin{tabular}[c]{@{}c@{}}Avg.\\ user\\ rank\end{tabular}} & \multicolumn{1}{c|}{\begin{tabular}[c]{@{}c@{}}Overall\\ rank\end{tabular}} \\ \hline
\multicolumn{1}{|l|}{LNSP matting}                            & \multicolumn{1}{c|}{4.8}                                                             & \multicolumn{1}{c|}{5.9}                                                          & \multicolumn{1}{c|}{10.1}                                                       & \multicolumn{1}{c|}{6.9}                                                    \\ \hline
\multicolumn{1}{|l|}{CCM}                                     & \multicolumn{1}{c|}{11.5}                                                          & \multicolumn{1}{c|}{8.5}                                                          & \multicolumn{1}{c|}{6.9}                                                        & \multicolumn{1}{c|}{9}                                                    \\ \hline
\rowcolor[HTML]{CBCEFB} 
\multicolumn{1}{|l|}{\cellcolor[HTML]{CBCEFB}Proposed method} & \multicolumn{1}{c|}{\cellcolor[HTML]{CBCEFB}10}                                  & \multicolumn{1}{c|}{\cellcolor[HTML]{CBCEFB}9}                                  & \multicolumn{1}{c|}{\cellcolor[HTML]{CBCEFB}8.9}                                & \multicolumn{1}{c|}{\cellcolor[HTML]{CBCEFB}9.3}                            \\ \hline
\multicolumn{1}{|l|}{Comprehensive sampling}                  & \multicolumn{1}{c|}{9.4}                                                           & \multicolumn{1}{c|}{9.6}                                                          & \multicolumn{1}{c|}{10.6}                                                       & \multicolumn{1}{c|}{9.9}                                                    \\ \hline
\multicolumn{1}{|l|}{SVR matting}                             & \multicolumn{1}{c|}{13.6}                                                          & \multicolumn{1}{c|}{8.8}                                                           & \multicolumn{1}{c|}{9}                                                        & \multicolumn{1}{c|}{10.5}                                                     \\ \hline
\multicolumn{1}{|l|}{Comprehensive Weighted C\&T}             & \multicolumn{1}{c|}{11}                                                          & \multicolumn{1}{c|}{11.4}                                                          & \multicolumn{1}{c|}{10.5}                                                          & \multicolumn{1}{c|}{11}                                                   \\ \hline
\multicolumn{1}{|l|}{\cellcolor[HTML]{C0C0C0}Sparse coded matting}                    & \multicolumn{1}{c|}{\cellcolor[HTML]{C0C0C0}14.9}                                                           & \multicolumn{1}{c|}{\cellcolor[HTML]{C0C0C0}13.8}                                                         & \multicolumn{1}{c|}{\cellcolor[HTML]{C0C0C0}10.5}                                                       & \multicolumn{1}{c|}{\cellcolor[HTML]{C0C0C0}13}                                                   \\ \hline
\multicolumn{1}{|l|}{Weighted Color \& Texture}               & \multicolumn{1}{c|}{12}                                                          & \multicolumn{1}{c|}{14.3}                                                         & \multicolumn{1}{c|}{13.1}                                                          & \multicolumn{1}{c|}{13.1}                                                   \\ \hline
\multicolumn{1}{|l|}{Global sampling matting}                 & \multicolumn{1}{c|}{9.8}                                                            & \multicolumn{1}{c|}{16.5}                                                           & \multicolumn{1}{c|}{14.4}                                                       & \multicolumn{1}{c|}{13.5}                                                   \\ \hline
\multicolumn{1}{|l|}{Iterative transductive matting}          & \multicolumn{1}{c|}{15.3}                                                           & \multicolumn{1}{c|}{12.8}                                                          & \multicolumn{1}{c|}{14.5}                                                        & \multicolumn{1}{c|}{14.2}                                                    \\ \hline
\end{tabular}
}

\end{minipage}\\
\\
\\
%\hspace*{3cm}
\begin{minipage}{0.5\textwidth}
\centering
\resizebox{\textwidth}{!}{%
\begin{tabular}{lcccc}
                                                              & \multicolumn{4}{c}{\textbf{Gradient error}}                                                                                                                                                                                                                                                                                                       \\ \hline
\multicolumn{1}{|l|}{Method}                                  & \multicolumn{1}{c|}{\begin{tabular}[c]{@{}c@{}}Avg. \\ small\\  rank\end{tabular}} & \multicolumn{1}{c|}{\begin{tabular}[c]{@{}c@{}}Avg.\\ large \\ rank\end{tabular}} & \multicolumn{1}{c|}{\begin{tabular}[c]{@{}c@{}}Avg.\\ user\\ rank\end{tabular}} & \multicolumn{1}{c|}{\begin{tabular}[c]{@{}c@{}}Overall\\ rank\end{tabular}} \\ \hline
\rowcolor[HTML]{CBCEFB} 
\multicolumn{1}{|l|}{\cellcolor[HTML]{CBCEFB}Proposed method} & \multicolumn{1}{c|}{\cellcolor[HTML]{CBCEFB}6.3}                                  & \multicolumn{1}{c|}{\cellcolor[HTML]{CBCEFB}6.1}                                  & \multicolumn{1}{c|}{\cellcolor[HTML]{CBCEFB}11.4}                                & \multicolumn{1}{c|}{\cellcolor[HTML]{CBCEFB}7.9}  
\\ \hline  
\multicolumn{1}{|l|}{LNSP matting}          & \multicolumn{1}{c|}{6.8}                                                           & \multicolumn{1}{c|}{7.6}                                                          & \multicolumn{1}{c|}{11.4}                                                        & \multicolumn{1}{c|}{8.6}                                                    \\ \hline
\multicolumn{1}{|l|}{Comprehensive sampling}                            & \multicolumn{1}{c|}{9}                                                             & \multicolumn{1}{c|}{8.1}                                                          & \multicolumn{1}{c|}{9.4}                                                       & \multicolumn{1}{c|}{8.8}                                                                            \\ \hline
\multicolumn{1}{|l|}{CCM}                  & \multicolumn{1}{c|}{12.8}                                                           & \multicolumn{1}{c|}{10}                                                          & \multicolumn{1}{c|}{8.9}                                                       & \multicolumn{1}{c|}{10.5}                                                    \\ \hline
\multicolumn{1}{|l|}{SVR matting}                             & \multicolumn{1}{c|}{12.9}                                                          & \multicolumn{1}{c|}{11.4}                                                           & \multicolumn{1}{c|}{7.6}                                                        & \multicolumn{1}{c|}{10.6}                                                     \\ \hline
\multicolumn{1}{|l|}{\cellcolor[HTML]{C0C0C0}Sparse coded matting}             & \multicolumn{1}{c|}{\cellcolor[HTML]{C0C0C0}12.6}                                                          & \multicolumn{1}{c|}{\cellcolor[HTML]{C0C0C0}9.8}                                                          & \multicolumn{1}{c|}{\cellcolor[HTML]{C0C0C0}10.3}                                                          & \multicolumn{1}{c|}{\cellcolor[HTML]{C0C0C0}11}                                                   \\ \hline
\multicolumn{1}{|l|}{Segmentation-based matting}               & \multicolumn{1}{c|}{14.4}                                                          & \multicolumn{1}{c|}{9.8}                                                         & \multicolumn{1}{c|}{10}                                                          & \multicolumn{1}{c|}{11.4}                                                   \\ \hline
\multicolumn{1}{|l|}{Global sampling matting}                    & \multicolumn{1}{c|}{11.8}                                                           & \multicolumn{1}{c|}{12.6}                                                         & \multicolumn{1}{c|}{10.8}                                                       & \multicolumn{1}{c|}{11.7}                                                   \\ \hline
\multicolumn{1}{|l|}{Shared Matting}                 & \multicolumn{1}{c|}{11.9}                                                            & \multicolumn{1}{c|}{12.8}                                                           & \multicolumn{1}{c|}{10.9}                                                       & \multicolumn{1}{c|}{11.8}                                                   \\ \hline
\multicolumn{1}{|l|}{Improved color matting matting}                                     & \multicolumn{1}{c|}{13.5}                                                          & \multicolumn{1}{c|}{12.5}                                                          & \multicolumn{1}{c|}{10.1}                                                        & \multicolumn{1}{c|}{12}                                                    \\ \hline
\end{tabular}
}
\end{minipage}
\label{tab:ranking}
\end{table}
\begin{table}[t]
\caption{Evaluation of effect of each step in our method using sum of absolute difference (SAD) error averaged over all test images for three types of trimaps.}
\begin{center}
\begin{tabular}{lccc}
\multicolumn{4}{l}{\hspace*{4.7cm}Trimap}                                                                                                                                                                            \\
\hline
Term                    & \begin{tabular}[c]{@{}c@{}}Small \\      rank\end{tabular} & \begin{tabular}[c]{@{}c@{}}Large\\      rank\end{tabular} & \begin{tabular}[c]{@{}c@{}}User \\      rank\end{tabular} \\ \hline
Universal set           & 26.6                                                       & 26.1                                                      & 27.9                                                      \\
Initial estimate        & 18.3                                                       & 15.1                                                      & 14.3                                                      \\
Laplacian refinement of~\cite{johnson2014sparse} & 13                                                         & 11                                                        & 7.9                                                       \\
Final estimate       & 8                                                          & 7.6                                                       & 7.3                                                       \\ \hline
\end{tabular}
\end{center}
\label{tab:msecomp}
\end{table}
 
 \begin{table}[t]
  \caption{Quantitative comparison of our method with compressive matting~\cite{yoon2012alpha} using mean squared error (MSE) measure.}
  \small
  \resizebox{0.49\textwidth}{!}{%
  \begin{tabular}{cccc}
  Image                 &          & Compressive matting~\cite{yoon2012alpha} & Proposed method                    \\ \hline
  \multirow{2}{*}{GT01} & Trimap 1 & $5.0\times 10^{-4}$ & $\mathbf{1.83\times 10^{-4}}$ \\
                        & Trimap 2 & $8.1\times 10^{-4}$ & $\mathbf{2.20\times 10^{-4}}$ \\
  \multirow{2}{*}{GT02} & Trimap 1 & $12.0\times 10^{-4}$ & $\mathbf{3.58\times 10^{-4}}$  \\
                        & Trimap 2 & $15.0\times 10^{-4}$ &  $\mathbf{4.75\times 10^{-4}}$  \\ \hline
  \end{tabular}
  }
  \label{tab:compressive}
  \end{table}
  \begin{table}[t!]
  \caption{Quantitative evaluation of the proposed sparse coding framework with other feature coding algorithms.}
  \resizebox{0.48\textwidth}{!}{%
  \begin{tabular}{l|c|c|c|c|}
  \cline{2-5}
                                                   & \multicolumn{2}{c|}{\textbf{MSE}}                                                                               & \multicolumn{2}{c|}{\textbf{SAD}}                                                                               \\ \hline
  \multicolumn{1}{|l|}{Feature coding method}      & \begin{tabular}[c]{@{}c@{}}Small\\ trimap\end{tabular} & \begin{tabular}[c]{@{}c@{}}Large\\ trimap\end{tabular} & \begin{tabular}[c]{@{}c@{}}Small\\ trimap\end{tabular} & \begin{tabular}[c]{@{}c@{}}Large\\ trimap\end{tabular} \\ \hline
  \multicolumn{1}{|l|}{LLC}                        & 0.0241                                                 & 0.0207                                                 & 5.968                                                  & 7.28                                                   \\ \hline
  \multicolumn{1}{|l|}{{\color[HTML]{000000} LSC}} & 0.0176                                                 & 0.0271                                                 & 5.547                                                  & 10.62                                                  \\ \hline
  \multicolumn{1}{|l|}{Proposed method}            & \textbf{0.0118}                                        & \textbf{0.0155}                                        & \textbf{4.071}                                         & \textbf{5.82}                                          \\ \hline
  \end{tabular}
  }
  \label{tab:feature_coding}
  \end{table}
\footnotetext[2]{As on July 30, 2015}
  
Fig.~\ref{fig:prepost} shows the visual quality of the mattes obtained without using pre and post-processing steps. As sampling methods use a pairwise approach and ignore correlation among neighbors during alpha estimation, many discontinuities are present in the initial mattes of ~\cite{shahrian2013improving,varnousfaderani2013weighted} (Fig.~\ref{fig:prepost}(b,c)). The use of multiple unpaired $F$ and $B$ to reconstruct the color at a pixel leads to a better initial estimate in the proposed method.  
\subsubsection{Quantitative evaluation}
A quantitative evaluation of the proposed method at the alpha matting website~\cite{alphawebsite} is shown in Table~\ref{tab:ranking}. 
The proposed method ranks first in sum of absolute difference (SAD) and gradient error, and third in mean squared error (MSE) when compared to the current state-of-the-art. Table~\ref{tab:ranking} shows the relative ranking of the top 10 matting algorithms\protect\footnotemark[2] using the three error measures. Cluster based sampling matting has not been published at the time of submission of the paper and is not used for the comparison. 
 The proposed method also considerably improves on the previous work~\cite{johnson2014sparse} in all the three metrics. This is attributed to the effect of utilizing feature-space neighbors along with spatial neighbors to optimize the matte.
   
\begin{figure}[t]
  {\centering
  \includegraphics[width=0.98\linewidth, clip=true, trim=0.5cm 7.95cm 2.6cm 1.9cm]{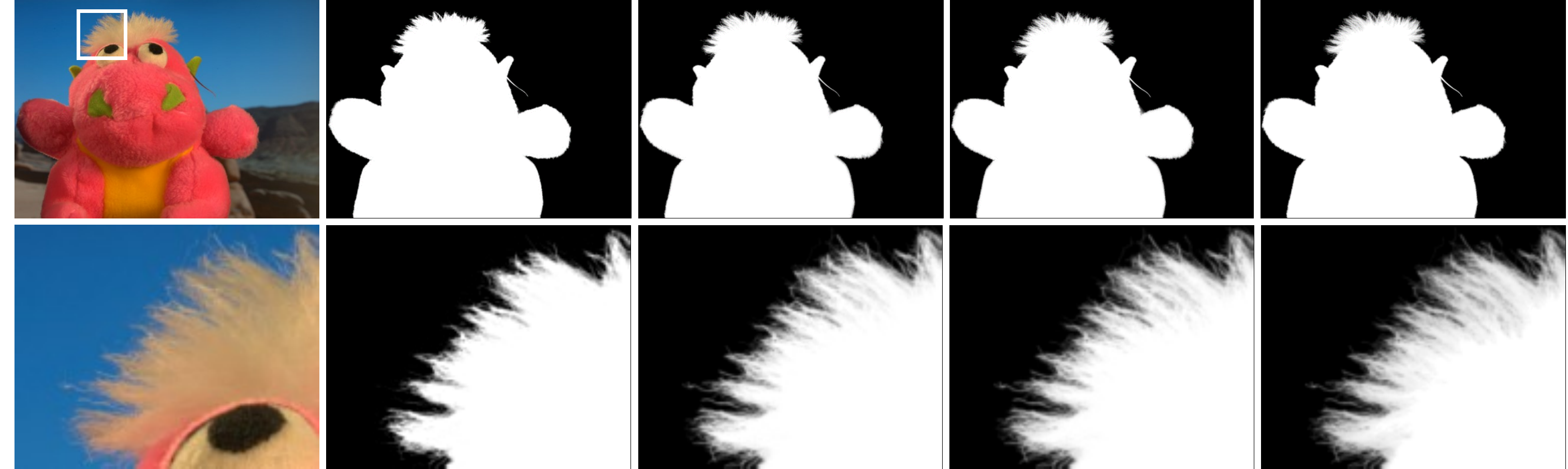}}
  \hspace*{0.7cm}(a)\hspace*{1.35cm}(b)\hspace*{1.35cm}(c)\hspace*{1.35cm}(d)\hspace*{1.35cm}(e)
  \caption{Visual comparison of feature coding methods for alpha matting. (a) Input image and zoomed window, (b) LLC , (c) LSC, (d) proposed method and (e) ground truth.}
  \label{fig:feature_coding}
  \end{figure}
  \begin{figure}[t]
  {\centering
  \includegraphics[width=0.98\linewidth, clip=true, trim=1.6cm 4.85cm 2.6cm 1.35cm]{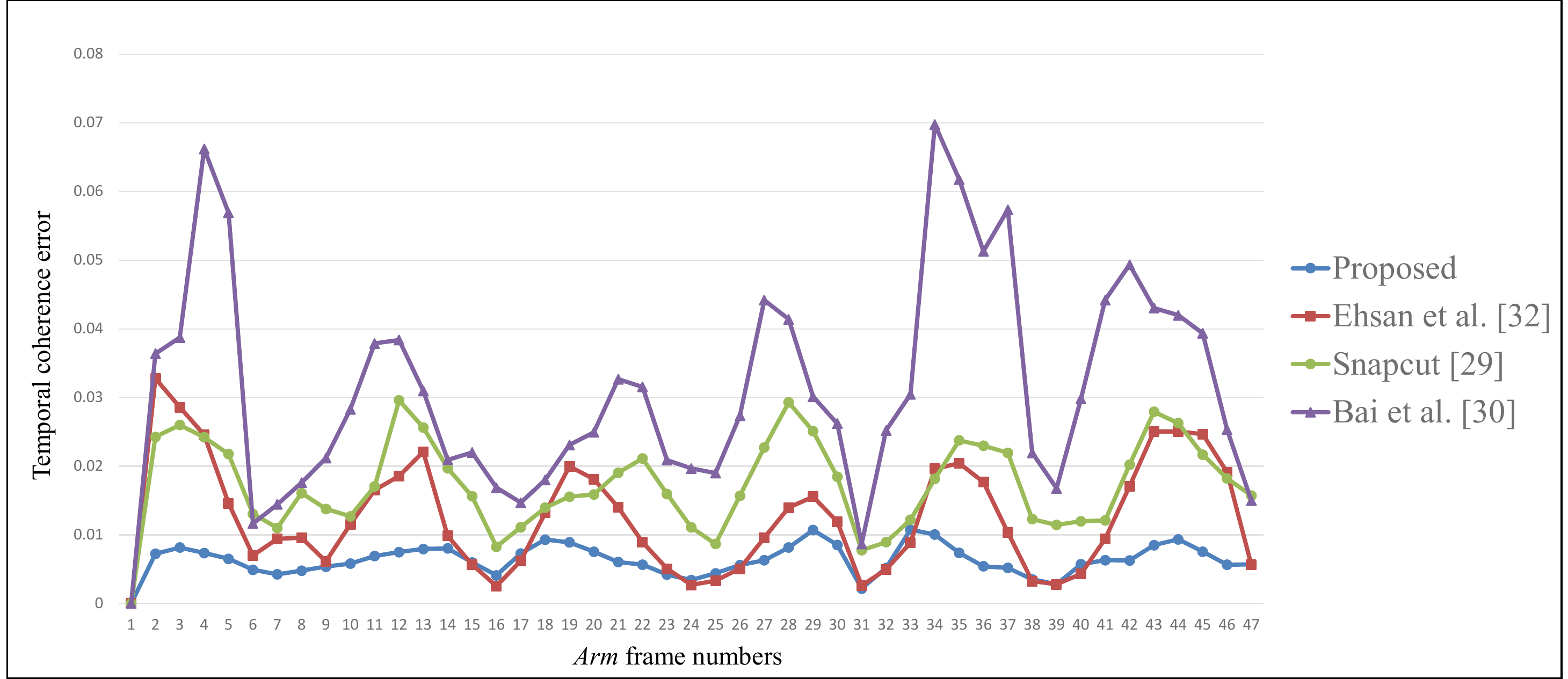}}
  \caption{Temporal coherence: Quantitative comparison with \cite{sharianvideo}, \cite{bai2009video} and \cite{bai2011towards} on the \textit{Arm} sequence. Our mattes not only give smaller error, but also maintains a stable temporal coherence across the frames.}
  \label{fig:tce_graph}
  \end{figure}
  \begin{table}[t!]
        \caption{Ranks of video matting methods with respect to mean squared error (MSE) and temporal coherence error (TCE) metrics.}
        \begin{minipage}{0.5\textwidth}
        \centering
        \resizebox{0.96\textwidth}{!}{%
        \begin{tabular}{lcccccc}
        \multicolumn{7}{c}{\hspace*{2cm}\textbf{MSE}}                                                                                                                                       \\ \hline
        \multicolumn{1}{|l|}{Method}   & Face          & Dancer       & Arm           & Smoke         & \multicolumn{1}{c|}{Woman}         & \multicolumn{1}{c|}{Overall}       \\ \hline
        \multicolumn{1}{|l|}{Proposed} & \textbf{1.39} & \textbf{1.0} & \textbf{1.40} & 3.0           & \multicolumn{1}{c|}{\textbf{1.20}} & \multicolumn{1}{c|}{\textbf{1.60}} \\
        \multicolumn{1}{|l|}{Ehsan \textit{et~al}.~\cite{sharianvideo}}    & 1.60          & 2.07         & 1.72          & \textbf{1.56} & \multicolumn{1}{c|}{1.79}          & \multicolumn{1}{c|}{1.75}          \\
        \multicolumn{1}{|l|}{Snapcut~\cite{bai2009video}}  & 3.02          & 3.02         & 3.03          & 1.77          & \multicolumn{1}{c|}{3.29}          & \multicolumn{1}{c|}{2.82}          \\
        \multicolumn{1}{|l|}{Bai \textit{et~al}.~\cite{bai2011towards}}      & 3.97          & 3.89         & 3.85          & 3.66          & \multicolumn{1}{c|}{3.70}          & \multicolumn{1}{c|}{3.81}          \\ \hline
        \end{tabular}
        }
        \end{minipage}\\
        \\
        \\
        \begin{minipage}{0.5\textwidth}
        \centering
        \resizebox{0.96\textwidth}{!}{%
        \begin{tabular}{lcccccc}
        \multicolumn{7}{c}{\hspace*{2.2cm}\textbf{TCE}}                                                                                                                                                \\ \hline
        \multicolumn{1}{|l|}{Method}           & Face          & Dancer        & Arm           & Smoke         & \multicolumn{1}{c|}{Woman}         & \multicolumn{1}{c|}{Overall}       \\ \hline
        \multicolumn{1}{|l|}{Proposed}         & 2.26          & \textbf{1.44} & \textbf{1.25} & 3.06          & \multicolumn{1}{c|}{\textbf{1.03}} & \multicolumn{1}{c|}{\textbf{1.81}} \\
        \multicolumn{1}{|l|}{Ehsan \textit{et al}.~\cite{sharianvideo}} & 2.52          & 1.60          & 1.95          & \textbf{1.16} & \multicolumn{1}{c|}{1.98}          & \multicolumn{1}{c|}{1.84}          \\
        \multicolumn{1}{|l|}{Snapcut~\cite{bai2009video}}     & \textbf{1.93} & 3.31          & 2.85          & 2.6           & \multicolumn{1}{c|}{3.55}          & \multicolumn{1}{c|}{2.85}          \\
        \multicolumn{1}{|l|}{Bai \textit{et al}.~\cite{bai2011towards}}  & 3.27          & 3.63          & 3.93          & 3.27          & \multicolumn{1}{c|}{3.43}          & \multicolumn{1}{c|}{3.5}           \\ \hline
        \end{tabular}
        }
        \end{minipage}
        \label{tab:video_mse_tce}
        \end{table}
 
 \begin{figure*}[t]
 \centering
 \includegraphics[width=0.92\linewidth, clip=true, trim=2.8cm 0cm 5.9cm 0cm]{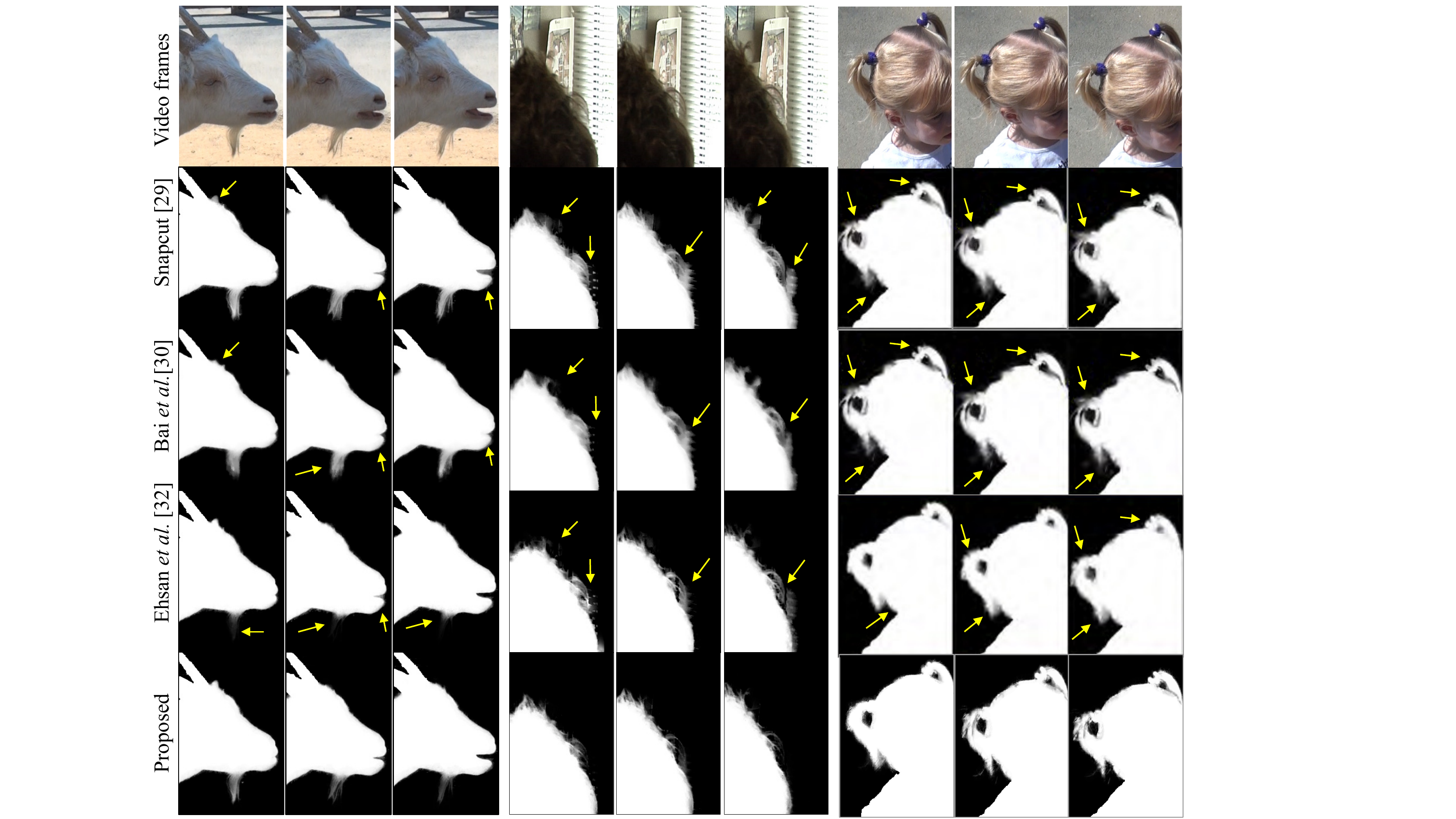}
 \caption[]{Qualitative comparison of the proposed method on \textit{Goat}, \textit{Cleanplate} and \textit{Girl} sequences respectively. Arrows indicate regions of low quality mattes, showing the effectiveness of our method. Dataset courtesy of Adobe.}
 \label{fig:video_qualitative}
 \end{figure*} 
Table~\ref{tab:msecomp} shows the contribution of each step of the proposed algorithm. The performance is measured using SAD error averaged over the test images, across three types of trimaps for each image - small, large and user. Universal set refers to the dictionary formed from the band of superpixels along the boundary of the unknown region. The high SAD error observed indicates that simply increasing the sample set for sparse coding does not result in better estimates due to the presence of high color-correlated samples to the unknown pixel at distances far from the immediate neighborhood. Initial estimate refers to the sparse coded estimate obtained on the refined dictionary, taking into account the certainty of the pixels. There is an improvement over the universal set because the spatial constraint controls the false correlated samples from being part of the sample set.
Laplacian refinement is the post-processing step which maintains the smoothness of the matte \cite{johnson2014sparse}. This stage shows a marked improvement over the sparse coding stage because in formulating the sparse code optimization, we do not consider the neighboring consistency. The final estimate obtained by our graph-based optimization using spatial and feature-space neighbors produces the lowest error rates across all sets of trimaps.
  
 Finally, we compare our results with the only other method that uses sparse coding in their formulation~\cite{yoon2012alpha}. As noted earlier, they do not provide results evaluated by \cite{alphawebsite} on the test images. Instead they provide quantitative evaluation on 2 training images (GT01,GT18), each on two trimaps, by giving the MSE that they obtained. We compared our results for the same images in Table~\ref{tab:compressive} where we show that we achieve a four-fold decrease in MSE on both the images.
\subsubsection{Comparison with other feature coding methods}
 The proposed method is compared with other feature coding algorithms to demonstrate their applicability to the problem of alpha matting. In particular, LSC~\cite{liu2011defense} and LLC~\cite{wang2010locality} are chosen and compared both qualitatively and quantitatively on the images. The parameters are not changed for this comparison. However, since LLC does not ensure that the returned feature codes are positive, the ratio of sum of squares of the $l_2$-norm of the codes are used to arrive at its alpha value. Fig.~\ref{fig:feature_coding} shows the visual quality of the mattes obtained using the 3 coding strategies. LLC biases the matte towards binary values while LSC produces more visually pleasing mattes. The use of specific constraints that steer the codes to alpha values in the proposed method produces the best visual results which agrees quantitatively as demonstrated by the error values presented in Table \ref{tab:feature_coding}.
     
\subsection{Video matting evaluation}
\begin{table*}[t]
  \centering
  \caption{Comparison of running time of the proposed method with recent sampling-based approaches.}
  \label{tab:timecomp}
  \begin{minipage}{0.58\linewidth}
  \centering
  \textbf{\small{Image matting}}\\
  \includegraphics[width=1\linewidth, clip=true, trim=2.4cm 19.55cm 2.52cm 2.5cm]{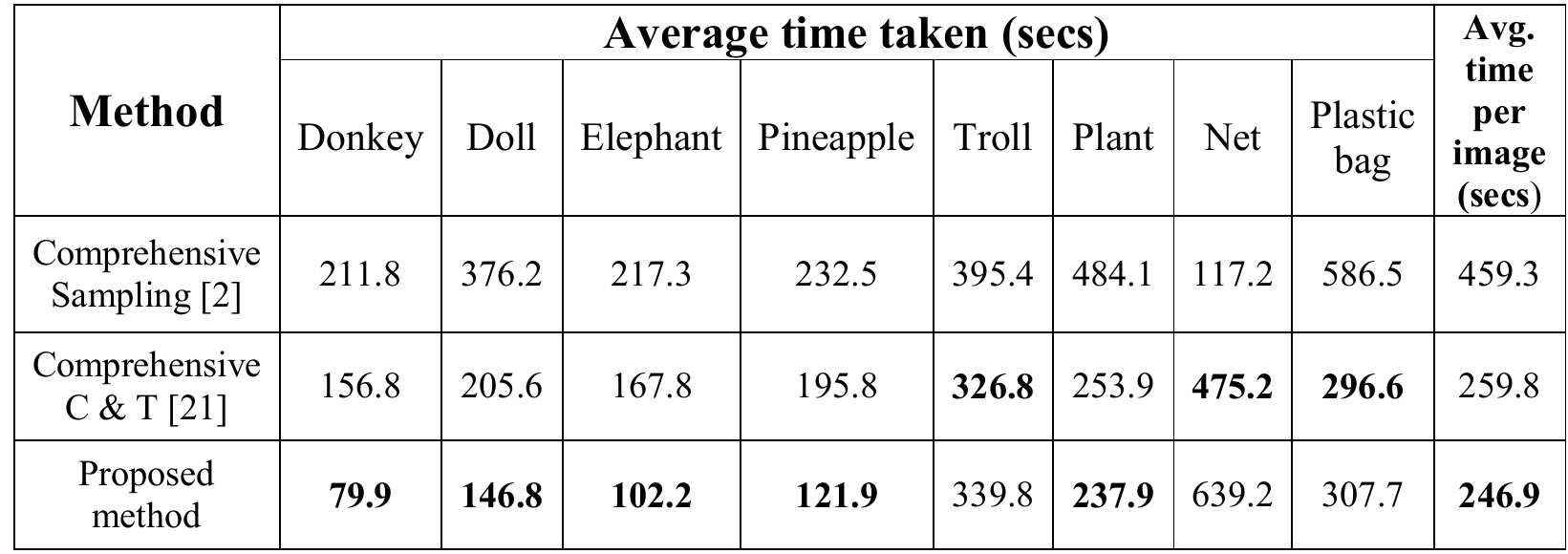}
  \end{minipage}
  \begin{minipage}{0.41\linewidth}
  \centering
  \textbf{\small{Video matting}}\\
  \vspace*{-0.1cm}
  \includegraphics[height=3.78cm, width=1\linewidth, clip=true, trim=2.4cm 10.45cm 2.52cm 9.7cm]{johnst6.pdf}
  \end{minipage}
  \end{table*} 
\subsubsection{Quantitative evaluation}
The dataset contains five videos with ground truth alpha available. We compare the proposed method against \cite{bai2011towards}, Snapcut~\cite{bai2009video} and \cite{sharianvideo}. These methods were designed to maintain temporal coherence in the mattes. First, the accuracy of the extracted mattes is measured by comparing against the ground truth using mean squared error criterion as shown in Table~\ref{tab:video_mse_tce}. The proposed method not only outperforms all the methods overall, but also on each video except one.

Temporal coherency is crucial for eliminating jitter and visually jarring artifacts in any video matting application. Being a subjective measure, it is not quite obvious as to what error metric can be used. Following \cite{sharianvideo}, temporal coherency is measured quantitatively as the mean squared error in the temporal derivative of the matte from the ground truth. Table~\ref{tab:video_mse_tce} shows the performance of the proposed video matting framework on the mean temporal coherence error metric over the five video sequences. The overall temporal coherence error is the least in the proposed method. Fig.~\ref{fig:tce_graph} shows the temporal coherence of our method on each frame of the \textit{Arm} sequence. As can be seen, our method produces the least fluctuations in temporal coherence while maintaining the lowest error possible. Additional comparisons are presented in the supplementary material.
\subsubsection{Qualitative evaluation}        
The performance of the proposed method is evaluated qualitatively on 3 sequences as shown in Fig.~\ref{fig:video_qualitative}. The low contrast between the goat's beard and the surroundings causes \cite{sharianvideo} to lose out the beard. The mattes obtained by \cite{bai2009video} and \cite{bai2011towards} oversmooth the matte resulting in spatial inaccuracy near the goat's mouth. We are able to achieve a good balance by picking the most relevant samples in building our dictionary. The cluttered background in the \textit{Cleanplate} sequence causes the matte to be inconsistent across consecutive frames, with background regions being misinterpreted as part of the foreground by considering spatial neighbors for smoothing. The proposed method is able to eliminate such inaccurate matching by removing the spatial neighbors and using only the feature-space neighbors for the graph model. Also, as can be seen from the \textit{Girl} sequence, the proposed approach is able to strike a good balance between temporal coherency and spatial accuracy by extracting mattes which are smoother in the temporal domain without losing out on the finer details like hair.    

We also provide qualitative comparison of the proposed method with the motion-aware KNN laplacian method~\cite{li2013motion} on the \emph{Amira} and \emph{Kim} sequences used in their paper in the supplementary video file. As the ground truth mattes for these sequences are not available, we do not present quantitative evaluations.
\subsection{Runtime performance}
Table \ref{tab:timecomp} compares the running time of the proposed method with recent sampling-based approaches. MATLAB implementations were evaluated on a PC running Intel Xeon 3.2 GHz processor. Sparse coding is able to generate faster estimates from the same set of $F$ and $B$ samples as opposed to an exhaustive pair search followed by conventional sampling methods~\cite{shahrian2013improving,varnousfaderani2013weighted} leading to better runtime performance. For the donkey image, the pre-processing and certainty estimation takes 32 seconds, the sparse coded matte estimation is completed in just 41 seconds followed by the graph-based optimization in 9 seconds.       
\paragraph*{Failure cases}
The proposed method fails to generate good quality mattes in the presence of illumination changes in highly transparent objects wherein the sparse codes are either biased to 0 or 1 for true mixed pixels. Examples are provided in the supplementary material.
\section{Conclusion}
\label{sec:conclusion}
In this paper, a new sampling based image matting method is presented that removes the restriction of ($F,B$) pairs in estimating the matte. In doing so, we are able to better approximate the matte that holds in simple separable regions as well as in complex textured regions that was verified by experimental evaluations and achieves state-of-the-art performance on the benchmark dataset. An extension to video is also presented which utilizes a graph model to encode the matte across a block of frames, resulting in better temporal coherency while maintaining the spatial accuracy of the matte in each frame. As with other sampling based approaches, the sparse codes are estimated locally at each pixel ignoring the global semantic information underlying in the image/video. The use of a low-rank approximation to the matting problem which enforces the global constraints is being investigated as future work.  

\ifCLASSOPTIONcaptionsoff
  \newpage
\fi



\bibliographystyle{IEEEtran}
\bibliography{IEEEabrv,IEEEexample}

\begin{thebibliography}{10}
\providecommand{\url}[1]{#1}
\csname url@samestyle\endcsname
\providecommand{\newblock}{\relax}
\providecommand{\bibinfo}[2]{#2}
\providecommand{\BIBentrySTDinterwordspacing}{\spaceskip=0pt\relax}
\providecommand{\BIBentryALTinterwordstretchfactor}{4}
\providecommand{\BIBentryALTinterwordspacing}{\spaceskip=\fontdimen2\font plus
\BIBentryALTinterwordstretchfactor\fontdimen3\font minus
  \fontdimen4\font\relax}
\providecommand{\BIBforeignlanguage}[2]{{%
\expandafter\ifx\csname l@#1\endcsname\relax
\typeout{** WARNING: IEEEtran.bst: No hyphenation pattern has been}%
\typeout{** loaded for the language `#1'. Using the pattern for}%
\typeout{** the default language instead.}%
\else
\language=\csname l@#1\endcsname
\fi
#2}}
\providecommand{\BIBdecl}{\relax}
\BIBdecl

\bibitem{shahrian2012weighted}
E.~Shahrian and D.~Rajan, ``Weighted color and texture sample selection for
  image matting,'' in \emph{Proc. IEEE CVPR}, 2012, pp. 718--725.

\bibitem{shahrian2013improving}
E.~Shahrian, D.~Rajan, B.~Price, and S.~Cohen, ``Improving image matting using
  comprehensive sampling sets,'' in \emph{Proc. IEEE CVPR}, 2013, pp. 636--643.

\bibitem{chen2013image}
X.~Chen, D.~Zou, S.~Z. Zhou, Q.~Zhao, and P.~Tan, ``Image matting with local
  and nonlocal smooth priors,'' in \emph{Proc. IEEE CVPR}, 2013, pp.
  1902--1907.

\bibitem{levin2008closed}
A.~Levin, D.~Lischinski, and Y.~Weiss, ``A closed-form solution to natural
  image matting,'' \emph{{IEEE} Trans. Pattern Anal. Mach. Intell.}, vol.~30,
  no.~2, pp. 228--242, 2008.

\bibitem{chen2012knn}
Q.~Chen, D.~Li, and C.-K. Tang, ``Knn matting,'' in \emph{Proc. IEEE CVPR},
  2012, pp. 869--876.

\bibitem{chuang2001bayesian}
Y.-Y. Chuang, B.~Curless, D.~H. Salesin, and R.~Szeliski, ``A bayesian approach
  to digital matting,'' in \emph{Proc. IEEE CVPR}, vol.~2, 2001, pp. II--264.

\bibitem{he2011global}
K.~He, C.~Rhemann, C.~Rother, X.~Tang, and J.~Sun, ``A global sampling method
  for alpha matting,'' in \emph{Proc. IEEE CVPR}, 2011, pp. 2049--2056.

\bibitem{gastal2010shared}
E.~S. Gastal and M.~M. Oliveira, ``Shared sampling for real-time alpha
  matting,'' in \emph{Computer Graphics Forum}, vol.~29, no.~2.\hskip 1em plus
  0.5em minus 0.4em\relax Wiley Online Library, 2010, pp. 575--584.

\bibitem{beiiterative2013}
B.~He, G.~Wang, C.~Shi, X.~Yin, B.~Liu, and X.~Lin, ``Iterative transductive
  learning for alpha matting,'' in \emph{Proc. IEEE ICIP}, Sept 2013, pp.
  4282--4286.

\bibitem{shi2013color}
Y.~Shi, O.~C. Au, J.~Pang, K.~Tang, W.~Sun, H.~Zhang, W.~Zhu, and L.~Jia,
  ``Color clustering matting,'' in \emph{Proc. IEEE ICME}, 2013, pp. 1--6.

\bibitem{wang2007optimized}
J.~Wang and M.~F. Cohen, ``Optimized color sampling for robust matting,'' in
  \emph{Proc. IEEE CVPR}, 2007, pp. 1--8.

\bibitem{lazebnik2006beyond}
S.~Lazebnik, C.~Schmid, and J.~Ponce, ``Beyond bags of features: Spatial
  pyramid matching for recognizing natural scene categories,'' in \emph{Proc.
  IEEE CVPR}, vol.~2, 2006, pp. 2169--2178.

\bibitem{liu2011defense}
L.~Liu, L.~Wang, and X.~Liu, ``In defense of soft-assignment coding,'' in
  \emph{Proc. IEEE ICCV}, 2011, pp. 2486--2493.

\bibitem{wright2010sparse}
J.~Wright, Y.~Ma, J.~Mairal, G.~Sapiro, T.~S. Huang, and S.~Yan, ``Sparse
  representation for computer vision and pattern recognition,'' \emph{Proc.
  IEEE}, vol.~98, no.~6, pp. 1031--1044, 2010.

\bibitem{wang2010locality}
J.~Wang, J.~Yang, K.~Yu, F.~Lv, T.~Huang, and Y.~Gong, ``Locality-constrained
  linear coding for image classification,'' in \emph{Proc. IEEE CVPR}, 2010,
  pp. 3360--3367.

\bibitem{wright2009robust}
J.~Wright, A.~Y. Yang, A.~Ganesh, S.~S. Sastry, and Y.~Ma, ``Robust face
  recognition via sparse representation,'' \emph{{IEEE} Trans. Pattern Anal.
  Mach. Intell.}, vol.~31, no.~2, pp. 210--227, 2009.

\bibitem{yang2010image}
J.~Yang, J.~Wright, T.~S. Huang, and Y.~Ma, ``Image super-resolution via sparse
  representation,'' \emph{{IEEE} Trans. Image Process.}, vol.~19, no.~11, pp.
  2861--2873, 2010.

\bibitem{yang2009linear}
J.~Yang, K.~Yu, Y.~Gong, and T.~Huang, ``Linear spatial pyramid matching using
  sparse coding for image classification,'' in \emph{Proc. IEEE CVPR}, 2009,
  pp. 1794--1801.

\bibitem{johnson2014sparse}
J.~Johnson, D.~Rajan, and H.~Cholakkal, ``Sparse codes as alpha matte,'' in
  \emph{Proc. BMVC}, 2014.

\bibitem{wang2005iterative}
J.~Wang and M.~F. Cohen, ``An iterative optimization approach for unified image
  segmentation and matting,'' in \emph{Proc. IEEE ICCV}, vol.~2, 2005, pp.
  936--943.

\bibitem{varnousfaderani2013weighted}
E.~S. Varnousfaderani and D.~Rajan, ``Weighted color and texture sample
  selection for image matting,'' \emph{{IEEE} Trans. Image Process.}, vol.~22,
  no.~11, pp. 4260--4270, 2013.

\bibitem{zhang2012learning}
Z.~Zhang, Q.~Zhu, and Y.~Xie, ``Learning based alpha matting using support
  vector regression,'' in \emph{Proc. IEEE ICIP}, 2012, pp. 2109--2112.

\bibitem{yoon2012alpha}
S.~M. Yoon and G.~Yoon, ``Alpha matting using compressive sensing,''
  \emph{Electron. Lett.}, vol.~48, no.~3, pp. 153--155, 2012.

\bibitem{alphawebsite}
\BIBentryALTinterwordspacing
(2009) Alpha matting evaluation website. [Online]. Available:
  \url{http://www.alphamatting.com}
\BIBentrySTDinterwordspacing

\bibitem{wang2008image}
J.~Wang and M.~F. Cohen, ``Image and video matting: A survey,'' \emph{Found.
  Trends. Comput. Graph. Vis.}, vol.~3, no.~2, pp. 97--175, Jan. 2007.

\bibitem{chuang2002video}
Y.-Y. Chuang, A.~Agarwala, B.~Curless, D.~H. Salesin, and R.~Szeliski, ``Video
  matting of complex scenes,'' \emph{ACM Trans. Graph.}, vol.~21, no.~3, pp.
  243--248, 2002.

\bibitem{wang2005interactive}
J.~Wang, P.~Bhat, R.~A. Colburn, M.~Agrawala, and M.~F. Cohen, ``Interactive
  video cutout,'' in \emph{ACM Trans. Graph.}, vol.~24, no.~3.\hskip 1em plus
  0.5em minus 0.4em\relax ACM, 2005, pp. 585--594.

\bibitem{li2005video}
Y.~Li, J.~Sun, and H.-Y. Shum, ``Video object cut and paste,'' in \emph{ACM.
  Trans. Graph.}, vol.~24, no.~3.\hskip 1em plus 0.5em minus 0.4em\relax ACM,
  2005, pp. 595--600.

\bibitem{bai2009video}
X.~Bai, J.~Wang, D.~Simons, and G.~Sapiro, ``Video snapcut: robust video object
  cutout using localized classifiers,'' in \emph{ACM Trans. Graph.}, vol.~28,
  no.~3.\hskip 1em plus 0.5em minus 0.4em\relax ACM, 2009, p.~70.

\bibitem{bai2011towards}
X.~Bai, J.~Wang, and D.~Simons, ``Towards temporally-coherent video matting,''
  in \emph{Computer Vision/Computer Graphics Collaboration Techniques}.\hskip
  1em plus 0.5em minus 0.4em\relax Springer, 2011, pp. 63--74.

\bibitem{lee2010temporally}
S.-Y. Lee, J.-C. Yoon, and I.-K. Lee, ``Temporally coherent video matting,''
  \emph{Graph. Models.}, vol.~72, no.~3, pp. 25--33, 2010.

\bibitem{sharianvideo}
E.~Shahrian, B.~Price, S.~Cohen, and D.~Rajan, ``Temporally coherent and
  spatially accurate video matting,'' \emph{Computer Graphics Forum}, vol.~33,
  no.~2, pp. 381--390, 2014.

\bibitem{li2013motion}
D.~Li, Q.~Chen, and C.-K. Tang, ``Motion-aware knn laplacian for video
  matting,'' in \emph{Proc. IEEE ICCV}, 2013, pp. 3599--3606.

\bibitem{achanta2012slic}
R.~Achanta, A.~Shaji, K.~Smith, A.~Lucchi, P.~Fua, and S.~Susstrunk, ``Slic
  superpixels compared to state-of-the-art superpixel methods,'' \emph{{IEEE}
  Trans. Pattern Anal. Mach. Intell.}, vol.~34, no.~11, pp. 2274--2282, 2012.

\bibitem{ju2013progressive}
J.~Ju, J.~Wang, Y.~Liu, H.~Wang, and Q.~Dai, ``A progressive tri-level
  segmentation approach for topology-change-aware video matting,'' in
  \emph{Computer Graphics Forum}, vol.~32, no.~7.\hskip 1em plus 0.5em minus
  0.4em\relax Wiley Online Library, 2013, pp. 245--253.

\bibitem{mairal2010online}
J.~Mairal, F.~Bach, J.~Ponce, and G.~Sapiro, ``Online learning for matrix
  factorization and sparse coding,'' \emph{J. Mach. Learn. Res.}, vol.~11, pp.
  19--60, 2010.

\bibitem{rhemann2009perceptually}
C.~Rhemann, C.~Rother, J.~Wang, M.~Gelautz, P.~Kohli, and P.~Rott, ``A
  perceptually motivated online benchmark for image matting,'' in \emph{Proc.
  IEEE CVPR}, 2009, pp. 1826--1833.

\end{thebibliography}
%\begin{thebibliography}{1}
%\bibliographystyle{IEEEtran}
%\bibliography{mybibfile}
%\end{thebibliography}


\begin{IEEEbiography}[{\includegraphics[width=1in,height=1.25in,clip,keepaspectratio]{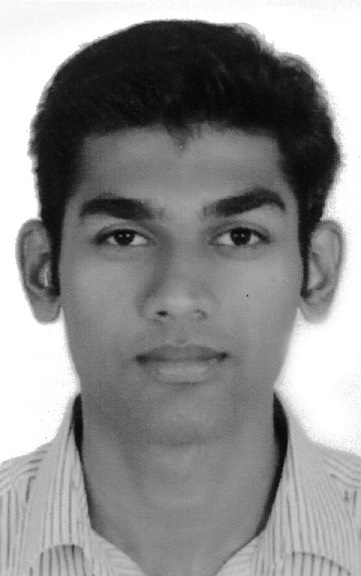}}]{Jubin Johnson}
received the Bachelor of Technology degree in electronics and communication engineering from Vellore Institute of Technology, India in 2010. From 2010 to 2012, he worked at Wipro Technologies, India. He is currently pursuing the PhD degree with the School of Computer Science and Engineering, Nanyang Technological University, Singapore. His current research interests include computer vision, image and video processing  and computer graphics.
\end{IEEEbiography}
\vfill

%\newpage
\begin{IEEEbiography}[{\includegraphics[width=1in,height=1.25in,clip,keepaspectratio]{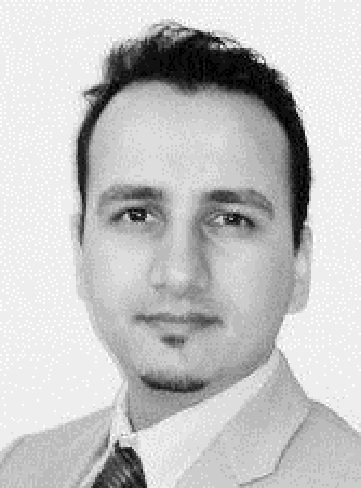}}]{Ehsan Shahrian Varnousfaderani}
received the B.Eng. (Hons.) degree in computer engineering (software engineering) from Karaj Islamic Azad University, Karaj, Iran, in 2006, the master’s degree in computer engineering - artificial intelligence and robotics from the Iran University of Science and Technology, Tehran, Iran, in 2008, and the Ph.D. degree from Nanyang Technological University, Singapore, in 2013. 
He was awarded the A*STAR Graduate Scholarship in 2008 and Adobe Research fund in 2012. He is currently a postdoctoral researcher of Shiley Eye Center at University of California San Diego.  His current research interests include computational ophthalmology, medical image analysis, computer graphics, image and video processing.
\end{IEEEbiography}
\vspace*{2.1cm}
\begin{IEEEbiography}[{\includegraphics[width=1in,height=1.25in,clip,keepaspectratio]{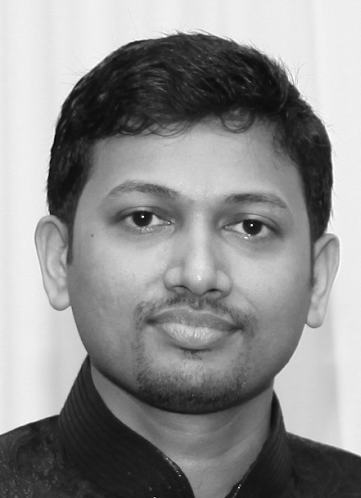}}]{Hisham Cholakkal}
received the Bachelor of Technology degree in electronics and communication engineering from Calicut University, India, in 2006 and the Master of Technology degree in signal processing from  Indian Institute of Technology Guwahati, India in 2009. From 2009 to 2012 he worked at Central Research Lab of Bharat Electronic Limited, Bangalore, India and  Advanced Digital Sciences Center, Singapore.  He is currently pursuing the PhD degree with the School of Computer Science and Engineering, Nanyang Technological University, Singapore.  His current research interests includes computer vision,  image processing, real-time video processing and machine learning.
\end{IEEEbiography}
\vspace*{2.1cm}
\begin{IEEEbiography}[{\includegraphics[width=1in,height=1.25in,clip,keepaspectratio]{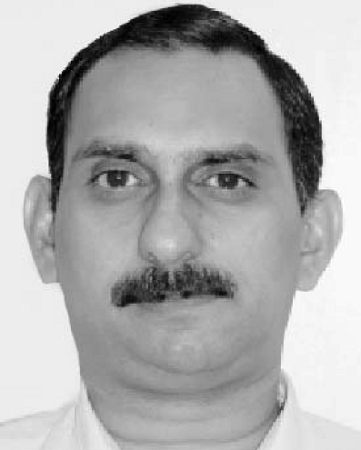}}]{Deepu Rajan}
received the Bachelor of Engineering
degree  in  electronics  and  communication  engineering  from  the  Birla  Institute  of  Technology,  Ranchi, India, the M.S. degree in electrical engineering from Clemson University, Clemson, SC, USA, and  the
Ph.D.  degree  from  the  Indian  Institute  of  Technology  Bombay,  Mumbai,  India.  He  is  an  Associate
Professor with the School of Computer Science and Engineering,
Nanyang Technological University, Singapore. From
1992 to 2002, he was a Lecturer with the Department
of  Electronics,  Cochin  University  of  Science  and
Technology,   Cochin,   India.
His  current   research   interests   include   image
processing,  computer  vision, and multimedia  signal processing.
\end{IEEEbiography}
\vfill

%\vfill


\end{document}